\RequirePackage[table]{xcolor}
\documentclass[journal,twoside,web]{ieeecolor}
\usepackage{tmi}
\usepackage{cite}
\usepackage{amsmath,amssymb,amsfonts}
\usepackage{algorithm}
\usepackage{algpseudocode}
\usepackage{float} 
\usepackage{graphicx}
\usepackage{textcomp}
\usepackage{multirow}
\usepackage{tikz}
\usepackage{bm}
\usepackage{booktabs}
\usepackage{pifont}
\usepackage{mathtools}

\usetikzlibrary{
patterns,calc,decorations.shapes,arrows,arrows.meta,positioning,fit,backgrounds,shadows.blur,shapes.geometric
}

\def\BibTeX{{\rm B\kern-.05em{\sc i\kern-.025em b}\kern-.08em
    T\kern-.1667em\lower.7ex\hbox{E}\kern-.125emX}}
\algrenewcommand\algorithmicindent{1ex}

\newcommand{\C}{\mathbb{C}}
\newcommand{\coils}{\mathbf{c}}
\newcommand{\kspace}{\mathbf{z}}
\newcommand{\motion}{\mathbf{v}}
\newcommand{\noise}{\boldsymbol{\epsilon}}
\newcommand{\etal}{\textit{et al}.}

\newcommand{\R}{\mathbb{R}}
\newcommand{\img}{\mathbf{x}}

\newcommand{\oW}{\mathcal{W}}
\newcommand{\oA}{\mathcal{A}}
\newcommand{\imgPrior}{\mathcal{R}_{\img}}
\newcommand{\motionPrior}{\mathcal{R}_{\motion}}
\newcommand{\coilsPrior}{\mathcal{R}_{\coils}}
\newcommand{\D}{\mathcal{D}}

\newcommand{\mF}{\mathbf{F}}
\newcommand{\mM}{\mathbf{M}}
\newcommand{\vm}{\mathbf{m}}

\newcommand{\prox}{\mathrm{prox}}
\newcommand{\diag}[1]{\mathrm{diag}(#1)}

\DeclareMathOperator*{\argmax}{argmax}
\DeclareMathOperator*{\argmin}{argmin}

\algnewcommand{\LineComment}[1]{\State \(\triangleright\) \textit{#1}}

\newcommand{\circlednum}[1]{\textcircled{\raisebox{-0.8pt}{\footnotesize#1}}}

\AtBeginEnvironment{figure}{\color{black}}
\AtBeginEnvironment{figure*}{\color{black}}
\AtBeginEnvironment{table}{\color{black}}
\AtBeginEnvironment{table*}{\color{black}}

\usepackage{caption}
\begin{document}
\title{MotionDPS: Motion-Compensated 3D Brain MRI Reconstruction}
\author{Antonio Ortiz-Gonzalez$^\ast$, Erich Kobler$^\ast$, Lukas Schletter and Alexander Effland
\thanks{AO is with the Life and Medical Sciences Institute, University of Bonn, Bonn, Germany (e-mail: aortizgo@uni-bonn.de).}
\thanks{EK is with the Institute for Machine Learning, LIT AI Lab, Department of Virtual Morphology, Clinical Research Institute Medical AI, Johannes Kepler University Linz, Linz, Austria (e-mail: erich.kobler@jku.at).}
\thanks{LS is with the
German Center for Neurodegenerative Diseases (DZNE), Bonn, Germany (e-mail: 
Lukas.Schletter@dzne.de).}
\thanks{AE is with the Institute for Applied Mathematics, University of Bonn, Bonn, Germany (e-mail: effland@iam.uni-bonn.de).}
\thanks{AO, LS, and AE acknowledge funding by the German Research Foundation (DFG) Grants EXC2151‐390873048 and EXC‐2047/1‐390685813.
EK acknowledges funding from the DFG within the SPP2298 under project 543939932 and from the Austrian Science Fund (FWF) project number 10.55776/COE12.}
\thanks{$^\ast$A. Ortiz-Gonzalez and E. Kobler share first authorship.}
}

\maketitle
\begin{abstract}

Magnetic resonance imaging (MRI) is highly susceptible to patient motion due to its relatively long acquisition times and the fact that data are acquired sequentially in k-space.
Even small patient movements introduce phase inconsistencies across measurements, leading to severe artifacts such as blurring, ghosting, and geometric distortions that can compromise diagnostic quality.
Retrospective motion compensation remains challenging, particularly in accelerated acquisitions, due to the ill-posed nature of the joint reconstruction and motion estimation problem.
In this work, we propose a unified Bayesian framework for motion-compensated 3D MRI that jointly estimates the anatomical image, rigid-body motion parameters, and coil sensitivity maps directly from motion-corrupted k-space data.
Our approach integrates pretrained 3D complex-valued score-based diffusion models as expressive anatomical image priors within a physics-based forward model.
Inference is performed by alternating diffusion posterior image updates with efficient proximal optimization steps for motion and coil sensitivity estimation, enabling fully unsupervised reconstruction without the need for paired motion-free training data.
Experiments on simulated and real-motion brain MRI datasets demonstrate that the proposed method achieves improved image quality and motion robustness compared to state-of-the-art classical and learning-based motion correction techniques, particularly in the presence of severe motion and high acceleration.

\end{abstract}


\begin{figure}[t]
\centering
\resizebox{\linewidth}{!}{
\begin{tikzpicture}[
  font=\small,
  text=black,
  panel/.style={inner sep=0pt, outer sep=0pt}
]

% ---- layout parameters ----
\def\W{2.0cm}      % width of each image
\def\H{2.3cm}      % height of each image
\def\xgap{0.03cm}   % horizontal gap between panels
\def\ygap{0.03cm}   % vertical gap between panels

% grid width (4 columns)
\pgfmathsetlengthmacro{\gridW}{4*\W + 3*\xgap}

% Column headers (top) --- axial row onlyx => row=1, header above at ~2*(H+ygap)
\node at ({0*(\W+\xgap)+0.5*\W}, {2.05*(\H+\ygap)}) {Reference};
\node at ({1*(\W+\xgap)+0.5*\W}, {2.05*(\H+\ygap)}) {RSS};
\node at ({2*(\W+\xgap)+0.5*\W}, {2.05*(\H+\ygap)}) {MotionTTT};
\node at ({3*(\W+\xgap)+0.5*\W}, {2.05*(\H+\ygap)}) {MotionDPS};

% Helper: place one image panel at (col,row)
\newcommand{\panelimg}[3]{% col, row, filename
  \node[panel, anchor=south west] at ({#1*(\W+\xgap)},{#2*(\H+\ygap)}) {%
    \includegraphics[width=\W,height=\H]{#3}%
  };
}

% ---------- Row 1: Axial ----------
\panelimg{0}{1}{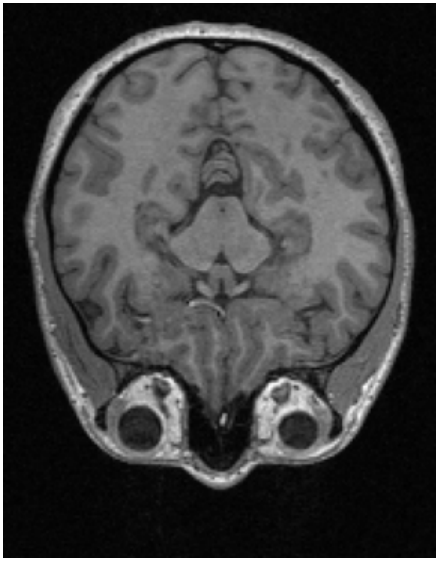}
\panelimg{1}{1}{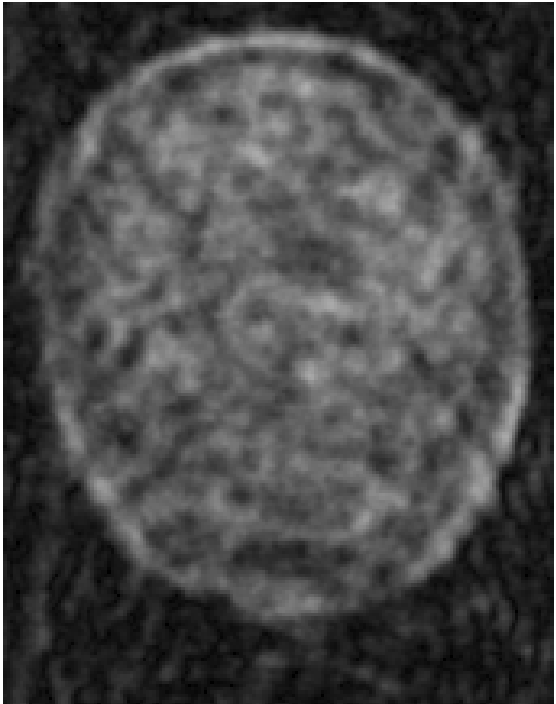}
\panelimg{2}{1}{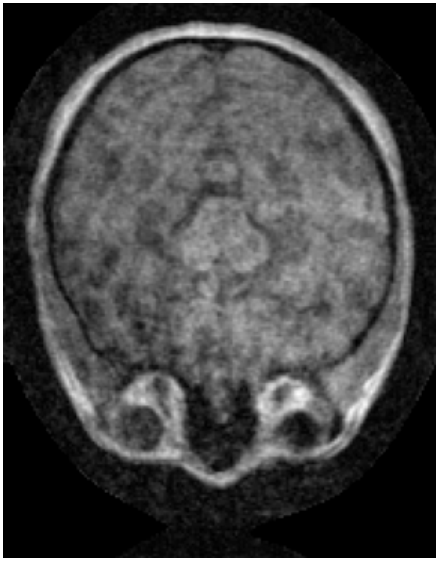}
\panelimg{3}{1}{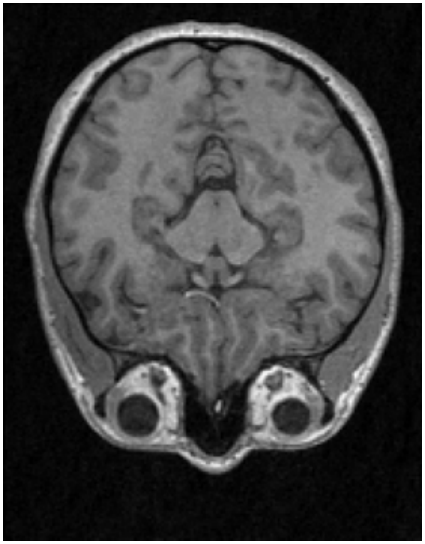}

% ---------- Row 0: Motion (directly under axial) ----------
\node[panel, anchor=north west] at (0,{0.95*(\H+\ygap)}) {%
  \includegraphics[width=\gridW]{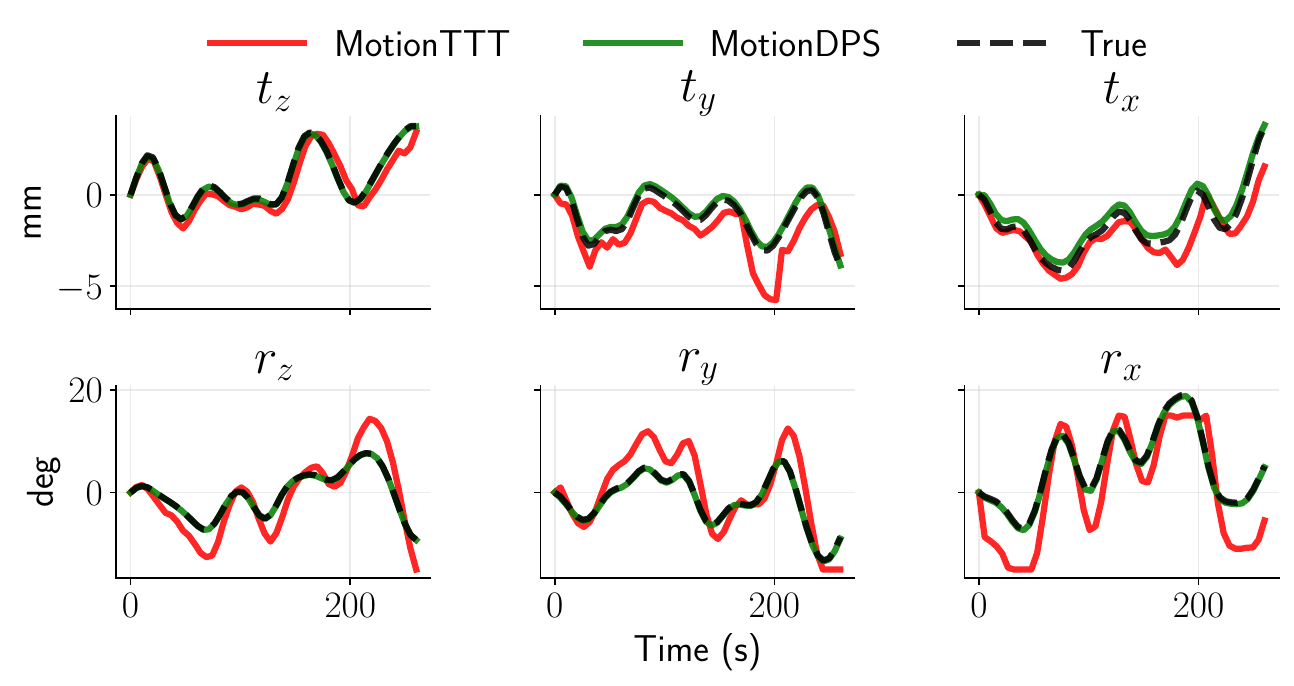}%
};

\end{tikzpicture}
}
\caption{Reconstruction results for a volume from the CC359 dataset under severe motion. 
Top pannel shows central axial views of the motion-free reference, root sum-of-squares (RSS), MotionTTT, and MotionDPS reconstruction.
MotionDPS substantially reduces motion artifacts compared to MotionTTT.
The bottom panel shows the estimated translational $(t_z,t_y,t_x)$ and rotational $(r_z,r_y,r_x)$ motion trajectories. 
}
\label{fig:cc359_recon_sub8}
\end{figure}

\section{Introduction}
\label{sec:introduction}

\IEEEPARstart{M}{agnetic} Resonance Imaging (MRI) is a versatile, non-invasive imaging modality that provides excellent soft-tissue contrast and enables the acquisition of rich anatomical and functional information.
However, its relatively long acquisition times render MRI highly susceptible to patient motion, which may arise from involuntary physiological processes (e.g., respiration, cardiac activity, brain pulsation) as well as voluntary actions such as head or eye movement and swallowing~\cite{Maclaren2013}.
Because MRI data are acquired in the frequency domain (k-space), motion introduces complex phase inconsistencies between successive signal measurements, leading to blurring, ghosting, geometric distortions, and reduced signal-to-noise ratio.
These artifacts degrade diagnostic confidence and necessitate repeated scans in up to 20\% of clinical examinations, resulting in substantial additional costs estimated at \$115{,}000 per scanner per year~\cite{Andre2015}.
In pediatric imaging, motion sensitivity further necessitates sedation or anesthesia as part of standard clinical care, increasing patient risk and incurring additional annual costs of up to \$319{,}000 per scanner~\cite{Slipsager2020, Eichhorn2021}.

Despite decades of research, motion correction (MC) remains a major unsolved challenge in clinical MRI.
Existing approaches are commonly categorized into \emph{prospective} and \emph{retrospective} techniques~\cite{Godenschweger2016, Slipsager2020}.
Prospective MC relies on specialized hardware to continuously track patient motion -- for example, using camera-based systems~\cite{Qin2009} -- and adapts the acquisition sequence in real time.
In contrast, retrospective MC operates solely on the corrupted k-space data after acquisition, avoiding additional hardware but requiring robust post-processing methods.

Classical retrospective MC methods combine explicit acquisition models with handcrafted regularization to address the ill-posedness of the reconstruction problem.
While the acquisition model enforces consistency between the estimated image, motion parameters, and measured k-space data, regularization is typically imposed via simple priors such as total generalized variation~\cite{knoll2011second} or entropy-based penalties~\cite{Loktyushin2013}.
Although effective in limited settings, these handcrafted priors struggle to capture the complexity and variability of real anatomical structures.
Recent advances in learning-based regularization have demonstrated that data-driven priors can model rich, non-linear image statistics and anatomical variability~\cite{kobler2020totaldeepvariationstable,narnhofer2021bayesianuncertaintyestimationlearned}.
When combined with physics-based data consistency, learned priors substantially improve robustness to undersampling, noise, and model mismatch by adaptively constraining the solution space of the MRI inverse problem~\cite{Heckel2024, Spieker2024}.

Beyond improving image regularization, these developments also motivate a more integrated treatment of the quantities involved in the MRI forward model.
The anatomical image, motion parameters, and coil sensitivity maps are tightly coupled: all three enter the acquisition process inseparably, and inaccuracies in any one propagate to the others.
In particular, motion corrupts the auto-calibration signal used for coil estimation, which in turn biases image reconstruction -- a cycle that decoupled or sequential pipelines cannot resolve.
Joint estimation therefore provides a principled framework for capturing these interdependencies and improving consistency across the reconstruction pipeline.

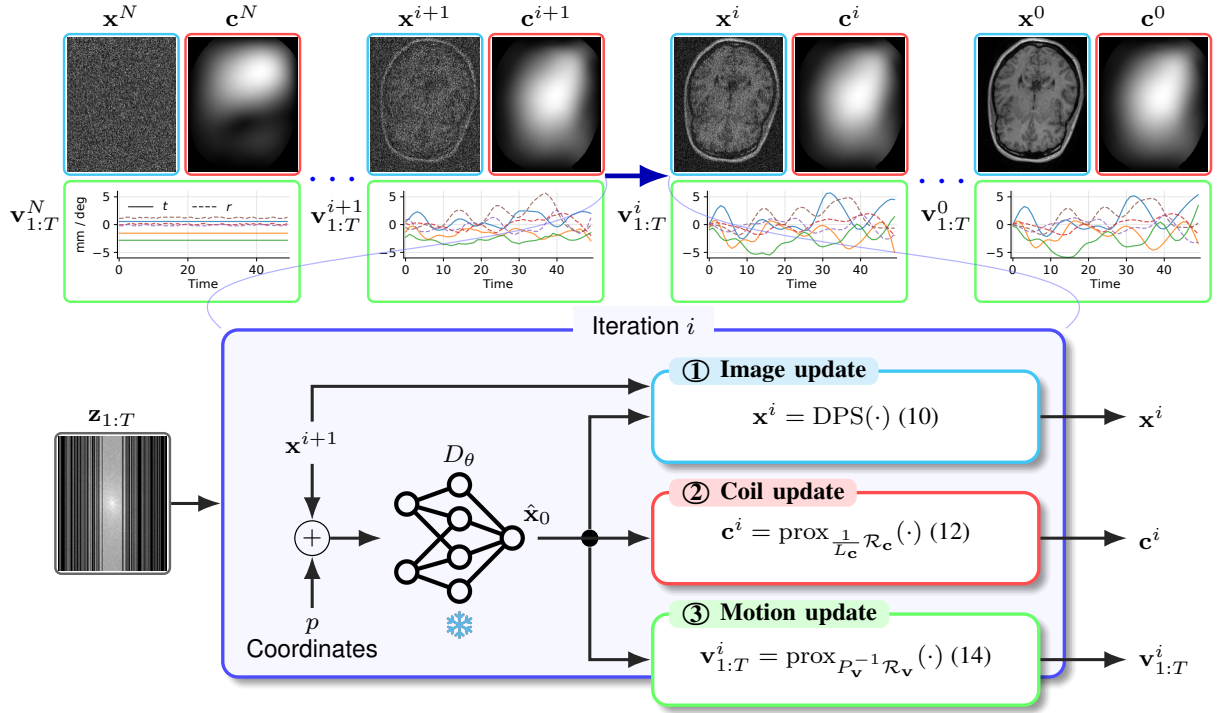
\begin{figure*}[t]
\centering
\resizebox{0.9\textwidth}{!}{
\begin{tikzpicture}[
  font=\sffamily\footnotesize,
  text=black,
  >=Latex,
  flow/.style={-Latex,line width=1.0pt,draw=black!85},
  blueflow/.style={-Latex,line width=1.55pt,draw=blue!70!black},
  block/.style={
    draw,
    rounded corners=5pt,
    line width=1.1pt,
    fill=white,
    blur shadow={shadow blur steps=4}
  },
  update/.style={
    block,
    minimum width=4.45cm,
    minimum height=1.05cm,
    align=center
  },
  title/.style={
    anchor=west,
    font=\sffamily\footnotesize,
    inner xsep=4pt,
    inner ysep=1pt
  }
]

% ================= LOCAL MACROS =================
\def\brainimg#1{\includegraphics[width=1.25cm,height=1.55cm]{#1}}
\def\snowflakeicon#1{\includegraphics[width=0.32cm]{#1}}
\def\neticon#1{\includegraphics[width=1.55cm]{#1}}
\def\motionimg#1{\includegraphics[width=2.64cm,height=1.32cm]{#1}}
% =================================================
% TOP DIFFUSION TRAJECTORY
% =================================================
\node[
    draw=cyan!60,
    line width=0.8pt,
    rounded corners=2pt,
    inner sep=1.0pt
] (xN)  at (0.2,3.8) {\brainimg{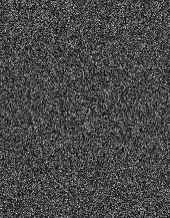}};
\node[
    draw=red!70,
    line width=0.8pt,
    rounded corners=2pt,
    inner sep=1.0pt
] (cN)  at (1.6,3.8) {\brainimg{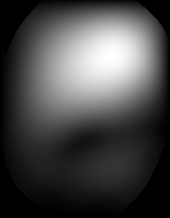}};
\node[
    draw=green!60,
    line width=0.8pt,
    rounded corners=2pt,
    inner sep=1.0pt
] (vN) at (0.9,2.22) {\motionimg{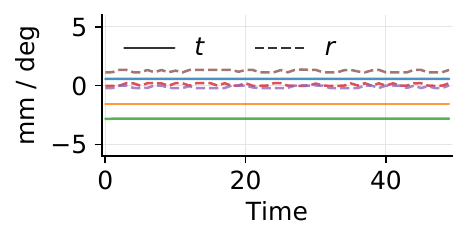}};

\node[
    draw=cyan!60,
    line width=0.8pt,
    rounded corners=2pt,
    inner sep=1.0pt
] (xi1) at (3.7,3.8) {\brainimg{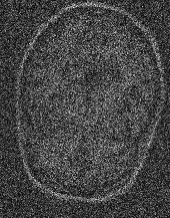}};
\node[
    draw=red!70,
    line width=0.8pt,
    rounded corners=2pt,
    inner sep=1.0pt
] (ci1) at (5.1,3.8) {\brainimg{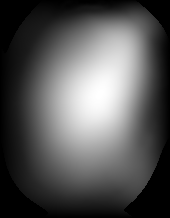}};
\node[
    draw=green!60,
    line width=0.8pt,
    rounded corners=2pt,
    inner sep=1.0pt
] (vi1) at (4.4,2.22) {\motionimg{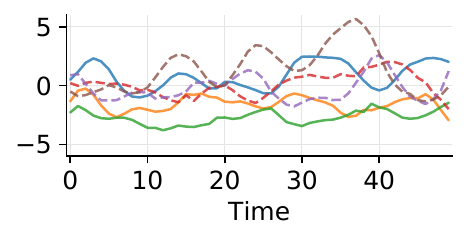}};

\node[
    draw=cyan!60,
    line width=0.8pt,
    rounded corners=2pt,
    inner sep=1.0pt
] (xi)  at (7.2,3.8) {\brainimg{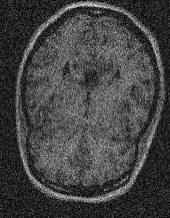}};
\node[
    draw=red!70,
    line width=0.8pt,
    rounded corners=2pt,
    inner sep=1.0pt
] (ci)  at (8.6,3.8) {\brainimg{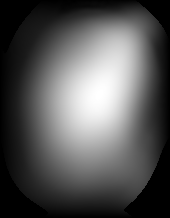}};
\node[
    draw=green!60,
    line width=0.8pt,
    rounded corners=2pt,
    inner sep=1.0pt
] (vi) at (7.9,2.22) {\motionimg{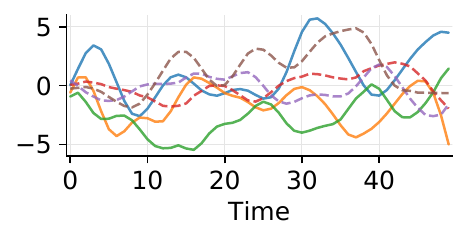}};

\node[
    draw=cyan!60,
    line width=0.8pt,
    rounded corners=2pt,
    inner sep=1.0pt
] (x0)  at (10.7,3.8) {\brainimg{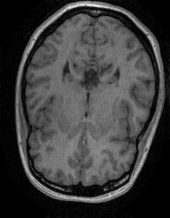}};
\node[
    draw=red!70,
    line width=0.8pt,
    rounded corners=2pt,
    inner sep=1.0pt,
] (c0)  at (12.1,3.8) {\brainimg{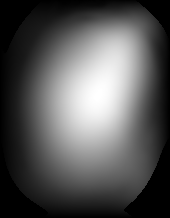}};
\node[
    draw=green!60,
    line width=0.8pt,
    rounded corners=2pt,
    inner sep=1.0pt
] (v0) at (11.4,2.22) {\motionimg{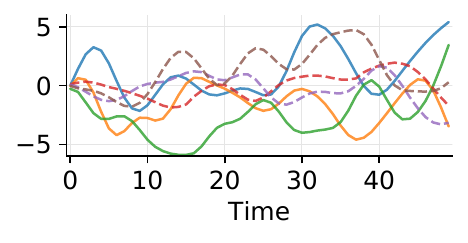}};

\node[above=-0.05cm of xN]  {$\img^{N}$};
\node[above=-0.05cm of xi1] {$\img^{i+1}$};
\node[above=-0.05cm of xi]  {$\img^{i}$};
\node[above=-0.05cm of x0]  {$\img^{0}$};
\node[above=-0.05cm of cN]  {$\coils^{N}$};
\node[above=-0.05cm of ci1] {$\coils^{i+1}$};
\node[above=-0.05cm of ci]  {$\coils^{i}$};
\node[above=-0.05cm of c0]  {$\coils^{0}$};
\node[left=-0.1cm of vN, yshift=3mm]  {$\motion^{N}_{1:T}$};
\node[left=-0.1cm of vi1, yshift=3mm]  {$\motion^{i+1}_{1:T}$};
\node[left=-0.1cm of vi, yshift=3mm]  {$\motion^{i}_{1:T}$};
\node[left=-0.1cm of v0, yshift=3mm]  {$\motion^{0}_{1:T}$};

\node[font=\Large, text=blue] at ($(xi1.south west)-(0.35,.05)$) {$\cdots$};
\draw[blueflow] (ci1.south east) -- (xi.south west);
\node[font=\Large, text=blue] at (9.7,2.9) {$\cdots$};

\node[
    draw=black!60,
    line width=0.8pt,
    rounded corners=2pt,
    inner sep=1.0pt,
] (kspace) at (0.1,-0.8)
{\brainimg{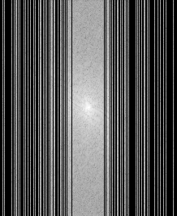}};

\node[above=-0.05cm of kspace]  {$\kspace_{1:T}$};

% =================================================
% ITERATION CONTAINER
% =================================================
\node[
  block,
  draw=blue!70,
  fill=blue!3,
  minimum width=9.70cm,
  minimum height=4.0cm,
  anchor=west
] (iter) at (1.35,-0.8) {};

\node[
  fill=blue!3,
  inner xsep=6pt
] at ($(iter.north)+(0,0.03)$)
{Iteration $i$};

% Left soft connector
\draw[
    blue!55,
    line width=0.4pt,
    % dashed,
    opacity=0.55
]
(ci1.south east)
.. controls +(0.4,-1.0) and +(-1.2,1.0) ..
(iter.north west);

% Right soft connector
\draw[
    blue!55,
    line width=0.4pt,
    % dashed,
    opacity=0.55
]
(xi.south west)
.. controls +(-0.4,-1.0) and +(1.2,1.0) ..
(iter.north east);

\draw[flow] (kspace.east) -- (iter.west);

% =================================================
% INTERNAL PIPELINE
% =================================================
\node (xin) at (2.40,-0.1) {$\img^{i+1}$};
\node[align=center] (coords) at (2.40,-2.35) {$p$\\Coordinates};

\node[circle,draw,minimum size=0.4cm,
  inner sep=0pt] (plus) at (2.40,-1.2) {$+$};

\draw[flow] (xin) -- (plus);
\draw[flow] (coords) -- (plus);

\node (net) at (4.10,-1.2)
{\neticon{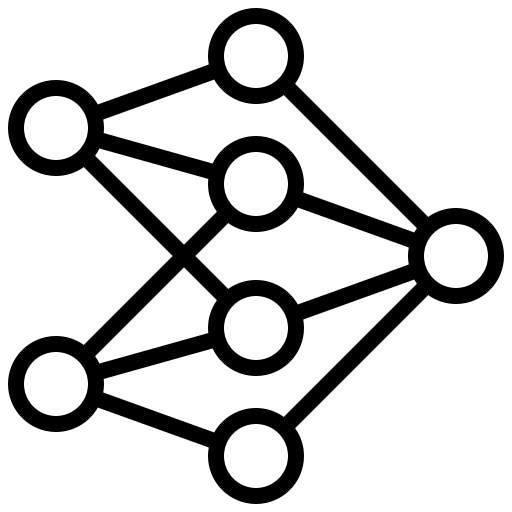}};

\node at ($(net)+(0,0.95)$) {$D_\theta$};
\node at ($(net)+(0,-1.0)$) {\snowflakeicon{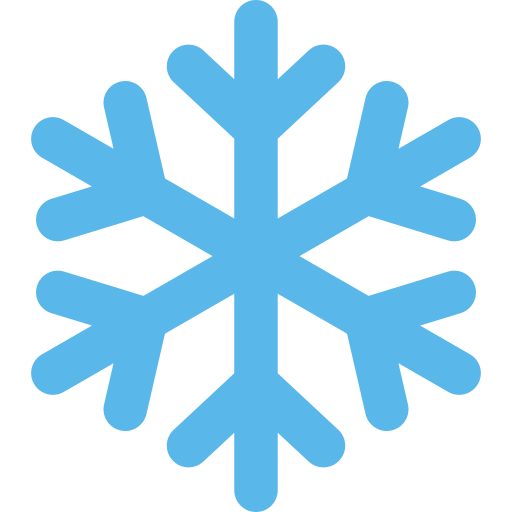}};

\draw[flow] (plus.east) -- (net.west);

\node[circle,fill=black,inner sep=2pt] (split) at (5.6,-1.2) {};
\node[above=1pt of net.east] {$\hat{\img}_0$};

% =================================================
% UPDATE BLOCKS - ALIGNED
% =================================================
\node[update,draw=cyan!60] (imgup)  at (8.55,0.2)
{$\img^{i}=\mathrm{DPS}(\cdot)$~\eqref{eq:dps_step}};

\node[update,draw=red!70] (coilup) at (8.55,-1.2)
{$\coils^{i}=\prox_{\frac{1}{L_\coils}\coilsPrior}(\cdot)$~\eqref{eq:prox_c}};

\node[update,draw=green!60] (motup) at (8.55,-2.6)
{$\motion^{i}_{1:T}=\prox_{P_\motion^{-1}\motionPrior}(\cdot)$~\eqref{eq:prox_v}};

\node[title,fill=cyan!15,font=\bfseries\footnotesize,rounded corners=3pt]  at ($(imgup.north west)+(0.20,-0.02)$)
{ \circlednum{1} Image update};

\node[title,fill=red!15,font=\bfseries\footnotesize,rounded corners=3pt]   at ($(coilup.north west)+(0.20,-0.02)$)
{\circlednum{2} Coil update};

\node[title,fill=green!15,font=\bfseries\footnotesize,rounded corners=3pt] at ($(motup.north west)+(0.20,-0.02)$)
{\circlednum{3} Motion update};

% split to updates
\draw[flow] (split) |- (imgup.west);
\draw[flow] (net.east) -- (coilup.west);
\draw[flow] (split) |- (motup.west);

% outputs
\draw[flow] (imgup.east) -- ++(1.0,0) node[right] {$\img^i$};
\draw[flow] (coilup.east) -- ++(1.0,0) node[right] {$\coils^i$};
\draw[flow] (motup.east) -- ++(1.0,0) node[right] {$\motion^i_{1:T}$};

\draw[flow] (xin) |- ($(imgup.west)+(0,0.35)$);

\end{tikzpicture}
}
\caption{Proposed MotionDPS framework for joint image reconstruction, coil sensitivity estimation, and motion estimation.}
\label{fig:motiondps_scheme}
\end{figure*}

In this work, we propose a unified framework for joint estimation of high-quality 3D images, motion parameters, and coil sensitivity maps from accelerated and potentially motion-corrupted MRI acquisitions.
Our method, termed \emph{MotionDPS}, integrates
\begin{itemize}
\item
score-based diffusion priors for high-fidelity anatomical regularization,
\item
a motion parameterization with temporal smoothness priors to enforce physiologically plausible trajectories, 
\item
a Tikhonov--Laplacian prior promoting spatially smooth coil sensitivity maps.
\end{itemize}
Inference proceeds by alternating posterior sampling via reverse-time diffusion for image updates with efficient proximal gradient updates for motion and coil estimation.
An overall schematic of the proposed method is shown in Fig.~\ref{fig:motiondps_scheme}.

This manuscript is organized as follows:
Section~\ref{sec:related_work} reviews related work on motion modeling, correction strategies, and learning-based MRI reconstruction.
Section~\ref{sec:methods} details the proposed MotionDPS framework.
Experimental results are presented in Section~\ref{sec:results}, followed by a discussion of implications and future research directions in Section~\ref{sec:discussion}.

\section{Related Work}
\label{sec:related_work}

Retrospective motion correction has traditionally been addressed using model-based reconstruction frameworks that jointly estimate rigid-body motion and the image through alternating reconstruction and registration steps, stabilized by handcrafted regularization~\cite{cordero2016, Johnson2019}.
These approaches explicitly enforce data consistency but tend to degrade under strong undersampling or complex motion patterns due to the limited expressiveness of the employed priors~\cite{kobler2020totaldeepvariationstable}.

More recently, learning-based motion correction methods have been proposed, most commonly relying on convolutional neural networks trained to map motion-corrupted reconstructions to artifact-free images.
The majority of these approaches operate purely in the image domain and are trained on synthetically corrupted data, which can limit generalization and may introduce hallucinated structures that violate the acquisition physics.
Furthermore, many learning-based methods are formulated in 2D or slice-wise hybrid 2D/3D settings, which restricts their applicability to fully 3D acquisitions where motion affects all spatial dimensions and strict k-space consistency is required~\cite{singh2022joint, kustner2019retrospective, Haskell2019, pawar2018MoCoNet, Hossbach2023, singh2023, levac2023, DIMA2025}.

Fully 3D motion compensation methods remain comparatively scarce but have gained increasing attention in recent years~\cite{cordero2016, Johnson2019, Duffy2021, almasni2021, klug2024}.
While these approaches better capture volumetric motion effects, joint estimation of anatomy, motion, and coil sensitivities in accelerated acquisitions remains largely underexplored.

Physics-informed deep learning approaches have emerged as a principled means of improving MRI reconstruction fidelity by explicitly incorporating the acquisition model into the learning process~\cite{hammernik2017}.
Building on this idea, learned image priors have emerged as powerful regularizers capable of modeling complex anatomical variability beyond handcrafted priors~\cite{Hemidi2025}. 
Among these, diffusion-based generative models have recently gained attention for MRI inverse problems.
Score-based diffusion models learn the gradient of the log-density of clean images and can be combined with the measurement likelihood via posterior sampling.
In particular, diffusion posterior sampling (DPS) augments the reverse diffusion process with data-consistency gradients, enabling reconstructions that remain faithful to both anatomy and acquisition data~\cite{chung2024, Valdivia2024}.

\subsection{Existing Approaches to Motion-Corrupted MRI Reconstruction}

Several recent works have extended diffusion-based methods to motion-corrupted MRI by incorporating motion into the forward model or using Bayesian score-based priors~\cite{chen2023jsmocojointcoilsensitivity, erlacher2023, Safari2024, Hemidi2025}.
However, most existing approaches address only part of the joint estimation problem, operate in 2D or simplified 3D settings, or rely on fixed coil sensitivity estimates, limiting their applicability to fully 3D accelerated acquisitions.
Table~\ref{tab:related_work_comparison} compares methods across eight key aspects of the joint reconstruction problem: fully 3D processing, complex-valued priors modeling both magnitude and phase, joint coil and motion estimation, motion regularization, k-space data consistency, diffusion priors, and unsupervised reconstruction (no paired motion-corrupted/motion-free training data required).
A method receives {\color{green!60!black}\ding{51}} if the 
capability is fully present, {\color{red}\ding{55}} if absent, and {\color{orange}$\sim$} if present but with significant limitations discussed in the text below.

\paragraph{Supervised and model-based methods}
Early retrospective motion-correction methods such as AltOpt~\cite{cordero2016} jointly estimate the image and motion within a physics-based reconstruction framework, but rely on handcrafted regularization and fixed pre-estimated coil sensitivities, making them sensitive to undersampling and motion-corrupted calibration.
More recent learning-based approaches employ deep neural networks for image-domain artifact correction.
Johnson~\etal~\cite{Johnson2019} use a conditional GAN for 3D motion correction without explicit data consistency or motion estimation, while Duffy~\etal~\cite{Duffy2021} similarly perform image-domain correction without motion or coil modeling.
Al-Masni~\etal~\cite{almasni2021} extend this direction using stacked U-Nets with neighboring-slice context, but still operate slice-wise without volumetric joint modeling, k-space consistency, or explicit motion and coil estimation.

\paragraph{Implicit prior/test-time training methods}
More recently, test-time optimization methods have incorporated pretrained networks as implicit image priors.
MotionTTT~\cite{klug2024} estimates 3D rigid motion through test-time gradient descent with k-space data consistency. 
However, its reconstruction network is trained with a magnitude-only loss and therefore does not model complex-valued MR signals (phase). 
In addition, coil sensitivities are estimated using ESPIRiT~\cite{Uecker2013}, limiting its applicability under realistic motion-corrupted acquisitions.

\begin{table*}[t]
\centering
\caption{Comparison of motion correction methods along key axes. 
{\color{green!60!black}\ding{51}}: fully supported; 
{\color{red}\ding{55}}: not supported; 
{\color{orange}$\sim$}: partially supported.}
\label{tab:related_work_comparison}
\resizebox{\textwidth}{!}{%
\begin{tabular}{lccccccccc}
\toprule
\textbf{Method} &
\textbf{3D} &
\textbf{Complex} &
\textbf{Joint coils} &
\textbf{Joint motion} &
\textbf{Motion reg.} &
\textbf{k-space} &
\textbf{Diffusion prior} &
\textbf{Unsupervised} \\
\midrule
AltOpt~\cite{cordero2016}
  & {\color{green!60!black}\ding{51}} & {\color{green!60!black}\ding{51}} & {\color{red}\ding{55}} & {\color{green!60!black}\ding{51}} & {\color{red}\ding{55}} & {\color{green!60!black}\ding{51}} & {\color{red}\ding{55}} & {\color{green!60!black}\ding{51}} \\
Johnson \etal~\cite{Johnson2019}
  & {\color{green!60!black}\ding{51}} & {\color{red}\ding{55}} & {\color{red}\ding{55}} & {\color{red}\ding{55}} & {\color{red}\ding{55}} & {\color{red}\ding{55}} & {\color{red}\ding{55}} & {\color{red}\ding{55}} \\
Duffy \etal~\cite{Duffy2021}
  & {\color{green!60!black}\ding{51}} & {\color{red}\ding{55}} & {\color{red}\ding{55}} & {\color{red}\ding{55}} & {\color{red}\ding{55}} & {\color{red}\ding{55}} & {\color{red}\ding{55}} & {\color{red}\ding{55}} \\
StackedUnet ~\cite{almasni2021}
  & {\color{orange}$\sim$} & {\color{red}\ding{55}} & {\color{red}\ding{55}} & {\color{red}\ding{55}} & {\color{red}\ding{55}} & {\color{red}\ding{55}} & {\color{red}\ding{55}} & {\color{red}\ding{55}} \\
MotionTTT~\cite{klug2024}
  & {\color{green!60!black}\ding{51}} & {\color{orange}$\sim$} & {\color{red}\ding{55}} & {\color{green!60!black}\ding{51}} & {\color{red}\ding{55}} & {\color{green!60!black}\ding{51}} & {\color{red}\ding{55}} & {\color{green!60!black}\ding{51}} \\
JSMoCo~\cite{chen2023jsmocojointcoilsensitivity}
  & {\color{orange}$\sim$} & {\color{red}\ding{55}} & {\color{orange}$\sim$} & {\color{orange}$\sim$} & {\color{red}\ding{55}} & {\color{green!60!black}\ding{51}} & {\color{green!60!black}\ding{51}} & {\color{green!60!black}\ding{51}} \\
Levac \etal~\cite{levac2023}
  & {\color{red}\ding{55}} & {\color{red}\ding{55}} & {\color{red}\ding{55}} & {\color{green!60!black}\ding{51}} & {\color{red}\ding{55}} & {\color{green!60!black}\ding{51}} & {\color{green!60!black}\ding{51}} & {\color{green!60!black}\ding{51}} \\
\textbf{MotionDPS (ours)}
  & {\color{green!60!black}\ding{51}} & {\color{green!60!black}\ding{51}} & {\color{green!60!black}\ding{51}} & {\color{green!60!black}\ding{51}} & {\color{green!60!black}\ding{51}} & {\color{green!60!black}\ding{51}} & {\color{green!60!black}\ding{51}} & {\color{green!60!black}\ding{51}} \\
\bottomrule
\end{tabular}%
}
\end{table*}

\paragraph{Diffusion-based methods}
Recent diffusion-based approaches integrate score-based priors into motion-corrupted MRI reconstruction.
JSMoCo~\cite{chen2023jsmocojointcoilsensitivity} jointly estimates motion and coil sensitivities within a score-based framework, but relies on a slice-wise formulation with a 3-DOF motion model and per-slice polynomial coil representations, limiting applicability to fully 3D accelerated acquisitions. 
In addition, motion is estimated independently per readout line without temporal regularization, resulting in unstable estimates and long runtimes due to annealed Langevin sampling.
Levac~\etal~\cite{levac2023} incorporate rigid motion into the reverse-diffusion likelihood in 2D, but assume fixed pre-estimated coil sensitivities, which remain sensitive to motion-corrupted k-space.

\subsection{Distinct Contributions of MotionDPS}
MotionDPS extends existing diffusion-based motion-correction methods by combining a fully 3D complex-valued diffusion prior with joint estimation of rigid-body motion and spatially smooth coil sensitivity maps within a single unsupervised framework.
Unlike prior approaches based on annealed Langevin dynamics or gradient-descent optimization, MotionDPS performs deterministic posterior inference through a single reverse-diffusion process with interleaved proximal updates and temporal motion regularization.
Experimentally, we evaluate the method on fully 3D accelerated acquisitions and include ablations isolating the impact of joint coil estimation and motion regularization (Section~\ref{sec:ablations}).

To the best of our knowledge, MotionDPS is the first fully 3D diffusion posterior sampling framework that jointly estimates the anatomical image, rigid-body motion, and coil sensitivity maps without requiring paired training data or pre-estimated coils.

\section{Methods}
\label{sec:methods}
Given motion-corrupted multi-coil k-space data, we aim to jointly estimate a high-quality image, coil sensitivity maps, and patient motion. We first model MRI acquisition, then present a joint reconstruction framework that combines coil/motion priors, pre-trained complex-valued 3D diffusion models, and physics-based forward models. Next, we describe the iterative algorithm (Alg.~\ref{alg:motiondps}) and its subproblems, and finally outline the diffusion-model training pipeline.

\subsection{Physics-based Imaging \& Motion Model}
In accelerated multi-coil MRI, the acquisition process is commonly modeled by the forward operator
\begin{align}
\kspace = \mM \mF (\coils \odot \img) + \noise,
\label{eq:forwardNoMotion}
\end{align}
where $\kspace\in\C^{C\times D}$ is the measured k-space data, $\img\in\C^{D}$ is the unknown image comprising of $D=r_z \times r_y \times r_x$ voxels, $\coils\in \C^{C\times D}$, the unknown spatially varying sensitivity maps of the $C$ receiver coils, and $\noise\in\C^{C\times D}$ is additive measurement noise.
The element-wise multiplication operator $\odot\colon \C^{C\times D}\times\C^D\to\C^{C\times D}$ is given by $\coils\odot\img=(\diag{\coils_1}\img,\ldots,\diag{\coils_C}\img)$.
$\mF\colon \C^{C\times D}\to\C^{C\times D}$ is the 3D discrete Fourier transform applied to each coil-weighted image in parallel.
Finally, $\mM$ is a block-diagonal matrix with $C$ identical binary acquisition masks $\diag{\vm}$ along the diagonal.
Here, $\vm\in\{0,1\}^D$ indicates if a particular k-space coefficient has been acquired.
To account for patient motion during acquisition, we explicitly model the sampling trajectory~$\mathcal{T}$, which specifies the k-space coefficients acquired at each discrete time point $t$.
We assume that $K$ coefficients are acquired simultaneously at any time point $t\in\{1,\ldots,T\}$, i.e. $\mathcal{T}\colon\{1,\ldots,T\}\to\{1,\ldots,D\}^K$.
Following \eqref{eq:forwardNoMotion}, the k-space data acquired at a single time point $\kspace_t\in\C^{C\times K}$ read as
\begin{equation}
\label{eq:forwardSingleTime}
\kspace_t = \mM_t\mF\left(\coils \odot \oW(\img, \motion_t)\right) + \noise_t = \oA_t(\img,\coils,\motion_t) + \noise_t,
\end{equation}
where $\motion_t\in\mathcal{V}$ is the unknown motion state at time $t$.
Here, a static reference image~$\img$ is transformed according to the motion state~$\motion_t$ via the nonlinear warping operator $\oW\colon\C^d\times \mathcal{V}\to\C^d$, where $\mathcal{V}$ denotes the space of valid motion states.
We implement $\oW$ using differentiable trilinear interpolation, ensuring well-defined gradients w.r.t. both image and motion states.
Then, the \emph{static} coil sensitivities~$\coils$ are applied, followed by the 3D discrete Fourier transform.
The binary masking operator~$\mM_t$ masks out all coefficients that are not acquired at time~$t$.
For notational brevity, we subsume all motion states into~$\motion_{1:T}\in\mathcal{V}^T$ and write
\begin{align}
\label{eq:forward}
\kspace_{1:T} = \oA(\img,\coils,\motion_{1:T}) + \noise.
\end{align}
Throughout this manuscript, we assume that the image degradation is caused by rigid motion, i.e., $\mathcal{V}=\mathrm{SE}(3)$, where $\mathrm{SE}(3)$ denotes the special Euclidean group with $\dim(\mathrm{SE}(3))=6$.
Note that the proposed approach can be easily extended to more complex and non-linear motion models by adapting $\mathcal{V}$.

As accelerated sampling trajectories~$\mathcal{T}$, we consider 3D Cartesian acquisition schemes with undersampling in the two phase-encoding dimensions and full sampling along the frequency-encoding (readout) direction, resulting in k-space lines.
Typically, multiple lines are acquired within a short time frame called a \emph{shot}, yielding $\kspace_s\in\C^{S}$ coefficients.
Acquisition is usually paused between shots to recover tissue magnetization, leading to longer delays between shots.
Thus, patient motion is more likely between shots than within a shot.
Two motion modeling paradigms are commonly considered: \emph{inter-shot motion} and \emph{intra-shot motion}~\cite{klug2024}.
Inter-shot motion assumes patients move only \emph{between} shots, whereas intra-shot motion also accounts for motion \emph{within} shots.
Note that the model in \eqref{eq:forward} can handle both sources.
For inter-shot motion, we set $K=S$ such that the number of time points equals the number of shots.
To model intra-shot motion, we define $n\geq 2$ and use $n$ times more motion states than shots, i.e. $nK=S$.

\subsection{Joint Reconstruction Framework}
To jointly reconstruct the high-quality image~$\img\in\C^D$, the coil sensitivities~$\coils\in\C^{C\times D}$, and all motion states $\motion_{1:T}\in\mathcal{V}^T$ from corrupted measurements $\kspace_{1:T}\in\C^{T\times C\times K}$, we adopt a Bayesian perspective.
To this end, we decompose the posterior probability into a likelihood and priors for the image~$p_X$, coils~$p_C$, and motion states~$p_V$ 
\begin{align*}
p(\img,\coils,\motion_{1:T}|\kspace_{1:T}) = p(\kspace_{1:T}|\img,\coils,\motion_{1:T}) p_X(\img) p_C(\coils) p_V(\motion_{1:T}).
\end{align*}
We estimate all unknowns via a maximum a posteriori (MAP) formulation,
$
(\hat{\img},\hat{\coils},\hat{\motion}_{1:T})\in\argmax_{\img,\coils,\motion_{1:T}}
p(\img,\motion,\coils|\kspace)$.
Taking the negative logarithm of the posterior yields the equivalent variational problem
\begin{align}
\label{eq:variational_obj}
(\hat{\img},\hat{\coils},\hat{\motion}_{1:T})\in&\argmin_{\img,\coils,\motion_{1:T}} \Big\{
\D\left(\oA\left(\img,\coils,\motion_{1:T}\right), \kspace_{1:T}\right) \\
&+ \imgPrior(\img)
+ \motionPrior(\motion)
+ \coilsPrior(\coils) \Big\},\notag
\end{align}
where the data fidelity term
\begin{equation}
\D\left(\oA(\img, \coils, \motion_{1:T}), \kspace_{1:T}\right)
= \tfrac{1}{2\sigma^2}
\|\oA(\img,\coils,\motion_{1:T}) - \kspace_{1:T}\|_2^2
\end{equation}
is associated with the likelihood of the measurements.
Here, we assume for simplicity additive Gaussian measurements noise~$\noise~\sim\mathcal{N}(0,\sigma^2\mathrm{Id})$.
Note that heteroscedastic noise could be easily integrated into this model.
In the negative log domain, the image~$\imgPrior$, coil~$\coilsPrior$, and motion state~$\motionPrior$ regularization models are associated with the corresponding prior.
In particular, we use an image prior implicitly given by
\begin{equation}
\label{eq:scoreDenoiser}
\nabla \log p_X(\img) = -\nabla \imgPrior(\img) \approx \frac{1}{\sigma^2} \left(D_\theta(\img,\sigma) - \img\right).
\end{equation}
This approximation corresponds to the Tweedie estimator under a variance-exploding diffusion process and has been widely adopted in diffusion posterior sampling frameworks.
In particular, $D_\theta\colon\C^D\times\R_+\to\C^D$ is a pre-trained \emph{complex-valued} denoiser of an associated diffusion model~\cite{Song2021} and $\sigma>0$ is a pre-defined noise level in the diffusion process.
To train the \emph{complex-valued} denoiser we follow the diffusion model of Karras \etal~\cite{Karras2022} as described in Section~\ref{sec:training_details}.

The coil sensitivity maps are typically very smooth.
Thus, we follow Zach \etal~\cite{Zach2023} and apply Tikhonov regularization for each coil sensitivity map, i.e.
\begin{equation}
\label{eq:coil_prior}
\coilsPrior(\coils) = \frac{\gamma}{2}\sum_{j=1}^C \left(\big\|\mathrm{D} \Re(\coils_j)\big\|_2^2 + \big\|\mathrm{D} \Im(\coils_j)\big\|_2^2 \right).
\end{equation}
Here, $\mathrm{D}\in\R^{3D\times D}$ denotes the discrete 3D spatial gradient operator and $\Re$ and $\Im$ denote the real and imaginary components of the complex-valued sensitivity maps.
The scalar weight~$\gamma>0$ balances the regularization strength.
To promote physically plausible motion trajectories, we impose a second-order temporal regularization that penalizes high curvature in the motion trajectory
\begin{equation}
\label{eq:motion_prior}
\motionPrior(\motion)
=
\frac{\eta}{2}
\sum_{t=2}^{T-1}
\big\|
\motion_{t+1} - 2\motion_t + \motion_{t-1}
\big\|_2^2,
\end{equation}
thereby discouraging abrupt changes while allowing smooth temporal variations.
Here $\eta>0$ controls the smoothness of the estimated motion trajectories.

\subsection{Diffusion Posterior Sampling for Motion Compensation}

We solve the joint reconstruction problem in \eqref{eq:variational_obj} using a hybrid optimization strategy that combines \emph{diffusion posterior sampling} (DPS)~\cite{chung2024} for image estimation with \emph{inertial proximal alternating linearized minimization} (iPALM)~\cite{Pock2016} for coil sensitivity and motion estimation.
The resulting algorithm, termed \emph{MotionDPS}, alternates between three coupled subproblems: (i) diffusion-guided image updates accounting for the posterior distribution conditioned on current coil sensitivities and motion states, (ii) proximal gradient updates for the coil sensitivity maps, and (iii) proximal gradient updates for the motion states.
This alternating structure enables the integration of a strong learned image prior with physics-based data consistency and explicit motion modeling.

Algorithm~\ref{alg:motiondps} summarizes the proposed MotionDPS framework.
Starting from an initial noise realization, coil sensitivity estimate, and zero motion, the algorithm iteratively progresses through a predefined diffusion noise schedule. At each diffusion step, the image is updated via a DPS step, followed by coil sensitivity and motion updates using proximal gradient steps derived from the data fidelity and corresponding priors. After completion of the diffusion process, the algorithm returns the final reconstructed image, coil sensitivity maps, and estimated motion trajectories.

\begin{algorithm}[t]
\caption{MotionDPS}
\label{alg:motiondps}
\begin{algorithmic}
\Require k-space data $\kspace_{1:T}$, noise schedule $\{\sigma^i\}_{i=N}^{0}$, pre-trained denoiser~$D_\theta$.

\LineComment{Initialization}
% \State Initialize $\img, \coils, \motion$ via~\eqref{eq:init}
\State $\img^N=\sigma^N\epsilon,\quad \epsilon\sim\mathcal{N}(0,\mathrm{Id})$
\State $\coils^N = \mF^{-1}(\kspace)/\sqrt{\sum_{c=1}^{C}|\mF^{-1}(\kspace_c)|^2}$
\State $\motion_{1:T}^N=0$

\For{$i = N-1,\ldots,0$}
    \State $\beta=\frac{N-i}{N-i+3}$ \Comment{Momentum}
    % \State $\img_0 \leftarrow D_\theta(\img,\sigma_i)$
    \LineComment{Image subproblem -- Section~\ref{sec:image_subproblem}}
    % \State $(\img, \img_0) \gets \text{EulerSolver}(D_\theta, \img, \sigma_i, \sigma_{i+1})$ 
    \State $\img^{i},\hat{\img}^0 = \text{DPS}(\img^{i+1},\coils^{i+1},\motion_{1:T}^{i+1},\kspace_{1:T},\sigma^i, D_\theta)$ 

    \LineComment{Coil subproblem -- Section~\ref{sec:coil_subproblem}}
    \State $\tilde{\coils} = \coils^{i+1} + \beta (\coils^{i+1} - \coils^{i+2})$
    \State $\coils^{i} = \prox_{\frac{1}{L_\coils}\coilsPrior}\left(\tilde{\coils} - \frac{1}{L_\coils}\nabla_\coils\mathcal{D}\left(\oA(\hat{\img}^0,\tilde{\coils},\motion_{1:T}^{i+1}),\kspace_{1:T}\right) \right)$

    \LineComment{Motion subproblem -- Section~\ref{sec:motion_subproblem}}
    \State $\tilde{\motion}_{1:T} = \motion_{1:T}^{i+1} + \beta (\motion_{1:T}^{i+1} - \motion_{1:T}^{i+2})$
    \State {\small $\motion_{1:T}^{i} =\prox_{P_\motion^{-1}\motionPrior}(\tilde{\motion}_{1:T}-P_\motion^{-1}\nabla_\motion\mathcal{D}\oA(\hat{\img}^{0},\coils^{i},\tilde{\motion}_{1:T}),\kspace_{1:T}))$}

\EndFor
\State \Return $(\img_0,\coils,\motion)$
\end{algorithmic}
\end{algorithm}

\subsubsection{Image Subproblem}
\label{sec:image_subproblem}

Given fixed coil sensitivities $\coils$ and motion states $\motion_{1:T}$, we update the image by sampling from the posterior $p(\img | \kspace_{1:T}, \coils, \motion_{1:T})$ using \emph{diffusion posterior sampling} (DPS)~\cite{chung2024}.
We adopt the probability flow ODE formulation of a variance-exploding diffusion process~\cite{Karras2022}, yielding
\begin{equation}
\label{eq:dps}
\tfrac{\mathrm{d}\img}{\mathrm{d}\hat{t}}
=
-\sigma(\hat{t})
\left(
\nabla_\img \log p(\img)
+
\nabla_\img \log p(\kspace_{1:T} \mid \img,\coils,\motion_{1:T})
\right)
\end{equation}
with $\hat{t}$ denoting diffusion time.
The score $\nabla_\img \log p(\img)$ is provided by a pretrained complex-valued denoiser \eqref{eq:scoreDenoiser}.
Following~\cite{chung2024}, the likelihood gradient is approximated as
\begin{equation*}
\nabla_\img \log p(\kspace_{1:T}|\img,\coils,\motion_{1:T})
\simeq
-\nabla_\img
\mathcal{D}\left(\oA(\hat{\img}_0,\coils,\motion_{1:T}),\kspace_{1:T}\right),
\end{equation*}
with $\hat{\img}_0 = D_\theta(\img,\sigma)$.
Using a first-order Euler discretization, the DPS update at diffusion step $i$ is
% \begin{align}
% \hat{\img}_0 &= D_\theta(\img^{i+1}, \sigma^i), \notag\\
% \img^i
% &=\img^{i+1} + \tfrac{\sigma^{i+1}-\sigma^i}{\sigma^i}(\img^{i+1}-\hat{\img}_0) \\
%  &\hspace*{3.4em}-\zeta^i \nabla_{\img} \mathcal{D}\left(\oA(\hat{\img}_0,\coils^i,\motion_{1:T}^i),\kspace_{1:T}\right),\notag
%  \label{eq:dps_step}
% \end{align}
\begin{equation}
\begin{aligned}
\hat{\img}_0 &= D_\theta(\img^{i+1}, \sigma^i), \\
\img^i
&=\img^{i+1} + \tfrac{\sigma^{i+1}-\sigma^i}{\sigma^i}(\img^{i+1}-\hat{\img}_0) \\
 &\hspace*{3.4em}-\zeta^i \nabla_{\img} \mathcal{D}\left(\oA(\hat{\img}_0,\coils^i,\motion_{1:T}^i),\kspace_{1:T}\right),
 \label{eq:dps_step}
\end{aligned}
\end{equation}
for $i=N-1,\ldots,0$.
The noise schedule $\{\sigma^i\}$ follows~\cite{Karras2022}
\[
\sigma^i
=
\left(
\sigma_{\max}^{1/\rho}
+
\tfrac{i}{N-1}
\left(
\sigma_{\min}^{1/\rho}
-
\sigma_{\max}^{1/\rho}
\right)
\right)^{\rho}
\]
with $\rho=7$ in all experiments; $\zeta^i$ implements a decreasing schedule to emphasize data consistency initially in the diffusion process.
Details on the training of $D_\theta$ are provided in Section~\ref{sec:training_details}.

\subsubsection{Coil Subproblem}
\label{sec:coil_subproblem}

With image $\img$ and motion states $\motion_{1:T}$ fixed, we update the coil sensitivities by minimizing a locally linearized approximation of the data fidelity term.
Let $\tilde{\coils}\in\mathbb{C}^{C\times D}$ be the current iterate and $L_\coils>0$ be the Lipschitz constant of the data fidelity gradient w.r.t.~$\coils$, we get
\begin{align}
\coils^i
=
\argmin_{\coils}
\Big\{
&\coilsPrior(\coils)
+
\frac{L_\coils}{2}\|\coils-\tilde{\coils}\|_2^2
+ \\
&\left\langle
\nabla_\coils
\mathcal{D}\!\left(\oA(\img,\tilde{\coils},\motion_{1:T}),\kspace_{1:T}\right),
\coils-\tilde{\coils}
\right\rangle
\Big\}. \notag
\end{align}
This problem admits the closed-form proximal gradient update
\begin{equation}
\coils^i
=
\prox_{\frac{1}{L_\coils}\coilsPrior}
\left(
\tilde{\coils}
-
\tfrac{1}{L_\coils}
\nabla_\coils
\mathcal{D}\left(\oA(\img,\tilde{\coils},\motion_{1:T}),\kspace_{1:T}\right)
\right).
\label{eq:prox_c}
\end{equation}
The proximal operator of the Tikhonov regularizer \eqref{eq:coil_prior} is solved in closed form using the discrete sine transform~\cite{Zach2023}.
Coil sensitivities are normalized as
\begin{equation}
    \coils = \frac{\coils}{\sqrt{\sum_{c}|\coils_c|^2}}
\end{equation}
in every step, and $L_\coils$ is locally estimated via backtracking.

\subsubsection{Motion Subproblem}
\label{sec:motion_subproblem}
Motion parameters are updated via preconditioned proximal gradient descent with preconditioning~$P_\motion$ (see below)
\begin{equation}
\begin{aligned}
\motion_{1:T}=&\prox_{P_\motion^{-1}\motionPrior}
(\tilde{\motion}_{1:T}-P_\motion^{-1}\nabla_{\motion}
\mathcal{D}(\oA(\img,\coils,\tilde{\motion}_{1:T}),\kspace_{1:T})).
\label{eq:prox_v}
\end{aligned}
\end{equation}
This proximal operator reduces to a linear system, which we solve using the conjugate gradient method.
Following Adabelief~\cite{Zhu2023}, we also introduce a diagonal motion preconditioner $P_\motion$ to reduce ill-conditioning of differing motion-parameter gradients
\begin{equation}
    P_\motion = L_\motion\mathrm{diag}\left( \sqrt{\frac{\bar{g}_{2}}{1-\beta_2^k}} + \epsilon\right),
\end{equation}
where $L_\motion$ denotes the backtracked local Lipschitz constant, $k$ the current iteration, $\epsilon>0$ is a small numerical stabilization constant, and $\bar{g}_2$ denotes the belief in the current gradient estimate.
Let $g=\nabla_\motion \mathcal{D}$, the running statistics are updated via
\begin{equation}
    \bar{g}_1 \gets \beta_1 \bar{g}_1 + (1-\beta_1)g, \quad \bar{g}_2 \gets \beta_2 \bar{g}_2 + (1-\beta_2)(g-\bar{g}_1)^2.
\end{equation}
We set the decay rates $\beta_1=0.5$ and $\beta_2=0.9$, and $\epsilon=10^{-8}$ in all experiments to accelerate convergence.

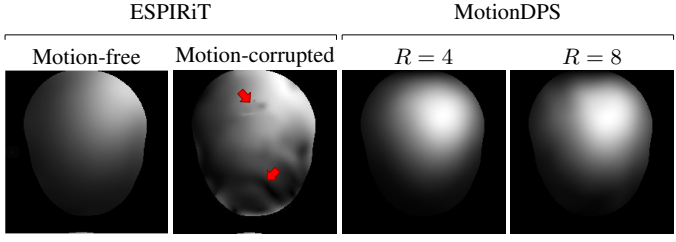
\begin{figure}[t]
\centering
\resizebox{\linewidth}{!}{\begin{tikzpicture}[
  font=\small,
  panel/.style={inner sep=0pt, outer sep=0pt},
  groupline/.style={line width=0.4pt},
]

% ---- layout parameters ----
\def\W{2.3cm}
\def\H{2.3cm}
\def\xgap{0.08cm}
\def\ygap{0.08cm}

% y positions
\pgfmathsetlengthmacro{\yCol}{2.05*(\H+\ygap)}     % column labels
\pgfmathsetlengthmacro{\yGroup}{2.32*(\H+\ygap)}   % group labels
\pgfmathsetlengthmacro{\yBracket}{2.18*(\H+\ygap)} % group rule

% ---- helper: image panel ----
\newcommand{\panelimg}[3]{% col, row, filename
  \node[panel, anchor=south west] at ({#1*(\W+\xgap)},{#2*(\H+\ygap)}) {%
    \includegraphics[width=\W,height=\H]{#3}%
  };
}

% ---- define coordinates for column centers ----
\coordinate (c0) at ({0*(\W+\xgap)+0.5*\W}, \yCol);
\coordinate (c1) at ({1*(\W+\xgap)+0.5*\W}, \yCol);
\coordinate (c2) at ({2*(\W+\xgap)+0.5*\W}, \yCol);
\coordinate (c3) at ({3*(\W+\xgap)+0.5*\W}, \yCol);

% group span edges
\coordinate (espL) at ({0*(\W+\xgap)}, \yBracket);
\coordinate (espR) at ({2*\W + 1*\xgap}, \yBracket);      % end of col 1
\coordinate (mdpsL) at ({2*(\W+\xgap)}, \yBracket);       % start of col 2
\coordinate (mdpsR) at ({4*\W + 3*\xgap}, \yBracket);     % end of col 3

% group label centers (midpoint between column centers, computed geometrically)
\path (c0) -- (c1) coordinate[pos=0.5] (espC);
\path (c2) -- (c3) coordinate[pos=0.5] (mdpsC);

% ---- group headers ----
\node[] at ($(espC)+(0,\yGroup-\yCol)$) {ESPIRiT};
\node[] at ($(mdpsC)+(0,\yGroup-\yCol)$) {MotionDPS};

% ---- group rules (brackets) ----
\draw[groupline] (espL) -- (espR);
\draw[groupline] (mdpsL) -- (mdpsR);
% small ticks (optional)
\draw[groupline] (espL) -- ++(0,-0.07cm);
\draw[groupline] (espR) -- ++(0,-0.07cm);
\draw[groupline] (mdpsL) -- ++(0,-0.07cm);
\draw[groupline] (mdpsR) -- ++(0,-0.07cm);

% ---- column labels ----
\node at (c0) {Motion-free};
\node at (c1) {Motion-corrupted};
\node at (c2) {$R=4$};
\node at (c3) {$R=8$};

% Row label (left, rotated)
% \node[rotate=90] at ({-0.2cm}, {1*(\H+\ygap)+0.5*\H}) {Axial};

% ---------- Row 1 ----------
\panelimg{0}{1}{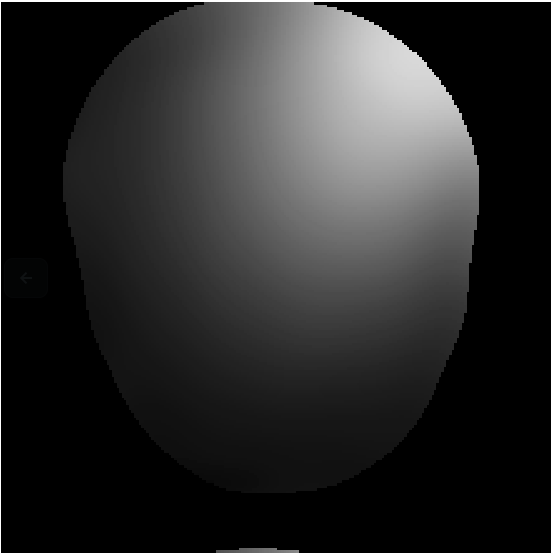}
\panelimg{1}{1}{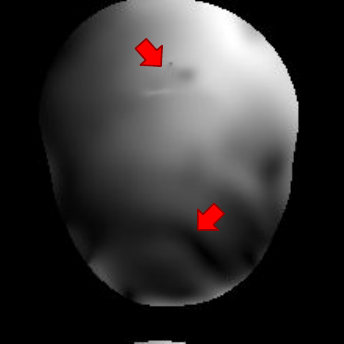}
\panelimg{2}{1}{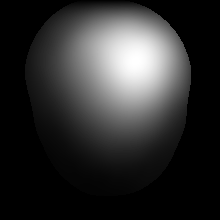}
\panelimg{3}{1}{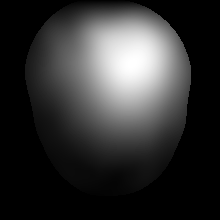}

\end{tikzpicture}}
\caption{Central coil sensitivity maps for a FastMRI sample. In the presence of motion, the ESPIRiT estimate exhibits visible motion-induced artifacts (red arrows). In contrast, MotionDPS yields smooth sensitivity maps closely resembling the motion-free ESPIRiT estimate. ESPIRiT sensitivity maps were computed with $R=4$.
}
\label{fig:fastmri_coils}
\end{figure}

\subsection{Training Complex-valued Denoising Diffusion Model}
\label{sec:training_details}

MotionDPS relies on a pretrained 3D complex-valued denoising diffusion model to provide score estimates during image updates.
Importantly, we employ a \emph{complex-valued} image prior to jointly model magnitude and phase information, which is essential for accurate motion-compensated reconstruction and for avoiding phase inconsistencies introduced by complex coil sensitivities and motion-induced k-space modulation.
Since the DPS update requires backpropagation through the denoiser, we adopt memory- and compute-efficient design choices to enable both training and inference on consumer-grade GPUs.
While the proposed framework is agnostic to the denoising architecture, we adapt the DRUNet~\cite{Zhang2022} model to 3D complex-valued data due to its favorable trade-off between reconstruction quality and computational efficiency.

The resulting network follows a U-Net-like encoder-decoder architecture with four resolution levels and residual blocks at each scale, and comprises approximately one million trainable parameters.
Conditioning on the noise level is achieved via a lightweight feed-forward embedding network, whose output modulates each residual block through feature-wise affine transformations, allowing $D_\theta$ to adapt across noise scales.

We employ a patch-based training strategy~\cite{Bieder2023}, where randomly sampled $128^3$ 3D patches are extracted from full MRI volumes.
To preserve global spatial coherence, each patch is augmented with its physical 3D RAS coordinates (in decimeters), which are concatenated channel-wise with the noisy input.
Formally, let $\img \in \mathbb{C}^D$ denote a clean image patch sampled from the data distribution $p_{\mathrm{data}}$, and let $\mathbf{p} \in \mathbb{R}^{3\times D}$ denote the corresponding spatial coordinates.
The denoiser is trained by minimizing
\begin{equation}
\mathbb{E}_{\img \sim p_{\mathrm{data}}}
\mathbb{E}_{\mathbf{n} \sim \mathcal{N}(0, \mathrm{Id})}
\left\|
D_\theta\left([\mathbf{z},\, \mathbf{p}], \sigma\right)
-
\img
\right\|_2^2,
\end{equation}
where $\mathbf{z} = \img + \sigma \mathbf{n}$ denotes the noisy input generated according to a variance-exploding diffusion process.
The noise level $\sigma$ is sampled from a log-normal distribution,
$\ln(\sigma) \sim \mathcal{N}(P_{\mathrm{mean}}, P_{\mathrm{std}}^2)$,
with $P_{\mathrm{mean}}=-1.2$ and $P_{\mathrm{std}}=1.2$.
We further adopt the EDM preconditioning scheme proposed by Karras \etal~\cite{Karras2022}.

\begin{table*}[t]
\caption{Quantitative results on FastMRI and CC359 under simulated motion.
Mean $\pm$ standard deviation for PSNR (dB) and SSIM are reported for mild, moderate, and severe motion at different acceleration factors $R$ ($R=1$ denotes fully sampled data).}
\centering
\renewcommand{\arraystretch}{1.5}
\resizebox{\linewidth}{!}{
\begin{tabular}{c|c|c|cc|cc|cc|cc}
\toprule[1.5pt]
\multirow{2}{*}{Dataset} & \multirow{2}{*}{R} & \multirow{2}{*}{\begin{tabular}[c]{@{}c@{}}Motion \\ severity\end{tabular}} & \multicolumn{2}{c|}{AltOpt} & \multicolumn{2}{c|}{MotionTTT} & \multicolumn{2}{c|}{StackedUnet} & \multicolumn{2}{c}{MotionDPS {\footnotesize(our)}} \\
 &  &  & PSNR & SSIM & PSNR & SSIM & PSNR & SSIM & PSNR & SSIM \\
\midrule[1pt]
\multirow{9}{*}{FastMRI} & \multirow{3}{*}{1} & Mild & 37.048 $\pm$ 3.331 & 0.937 $\pm$ 0.110 & 41.903 $\pm$ 2.158 & 0.976 $\pm$ 0.006 & 34.765 $\pm$ 3.028 & 0.954 $\pm$ 0.017 & \textbf{43.152 $\pm$ 1.672} & \textbf{0.981 $\pm$ 0.008} \\
 &  & Moderate & 32.201 $\pm$ 3.128 & 0.843 $\pm$ 0.221 & 36.542 $\pm$ 2.148 & 0.936 $\pm$ 0.154 & 30.441 $\pm$ 3.465 & 0.921 $\pm$ 0.034 & \textbf{40.560 $\pm$ 3.222} & \textbf{0.976 $\pm$ 0.012} \\
 &  & Severe & 24.075 $\pm$ 2.587 & 0.680 $\pm$ 0.138 & 24.154 $\pm$ 4.678 & 0.687 $\pm$ 0.135 & 29.144 $\pm$ 3.285 & 0.904 $\pm$ 0.046 & \textbf{32.210 $\pm$ 1.250} & \textbf{0.922 $\pm$ 0.013} \\
% \cline{2-11}
\cmidrule(lr){2-11}
 & \multirow{3}{*}{4} & Mild & 36.489 $\pm$ 3.038 & 0.928 $\pm$ 0.095  & \textbf{39.943 $\pm$ 1.759} & \textbf{0.969 $\pm$ 0.026} & 29.748 $\pm$ 2.173 & 0.909 $\pm$ 0.025 & 37.246 $\pm$ 0.767 & 0.967 $\pm$ 0.005 \\
 &  & Moderate & 29.147 $\pm$ 2.655 & 0.839 $\pm$ 0.223 & 34.762 $\pm$ 1.248 & 0.945 $\pm$ 0.047 & 27.453 $\pm$ 2.220 & 0.865 $\pm$ 0.043 & \textbf{35.938 $\pm$ 0.927} & \textbf{0.957 $\pm$ 0.008} \\
 &  & Severe & 24.937 $\pm$ 2.603 & 0.733 $\pm$ 0.069 & 26.720 $\pm$ 4.284 & 0.741 $\pm$ 0.165 & 25.455 $\pm$ 1.971 & 0.811 $\pm$ 0.054 & \textbf{33.176 $\pm$ 1.139} & \textbf{0.917 $\pm$ 0.017} \\
\cmidrule(lr){2-11}
% \cline{2-11}
 & \multirow{3}{*}{8} & Mild & 33.004 $\pm$ 3.068 & 0.913 $\pm$ 0.231 & 34.168 $\pm$ 1.155 & 0.918 $\pm$ 0.051 & 26.759 $\pm$ 1.258 & 0.864 $\pm$ 0.034 & \textbf{34.382 $\pm$ 0.940} & \textbf{0.939 $\pm$ 0.007} \\
 &  & Moderate & 28.662 $\pm$ 2.712 & 0.757 $\pm$ 0.173 & \textbf{34.862 $\pm$ 1.295} & 0.904 $\pm$ 0.066 & 26.085 $\pm$ 1.643 & 0.839 $\pm$ 0.055 & 33.120 $\pm$ 1.483 & \textbf{0.929 $\pm$ 0.012} \\
 &  & Severe & 23.354 $\pm$ 3.606 & 0.707 $\pm$ 0.137 & 26.704 $\pm$ 3.247 & 0.773 $\pm$ 0.061 & 25.126 $\pm$ 1.157 & 0.799 $\pm$ 0.049 & \textbf{31.172 $\pm$ 1.241} & \textbf{0.888 $\pm$ 0.013} \\
\midrule[1pt]
\multirow{6}{*}{CC359} & \multirow{3}{*}{1} & Mild & 39.140 $\pm$ 0.354 & \textbf{0.968 $\pm$ 0.003} & \textbf{39.327 $\pm$ 0.344} & 0.967 $\pm$ 0.030 & 33.068 $\pm$ 0.681 &  0.934 $\pm$ 0.008 & 36.952 $\pm$ 0.882 & 0.959 $\pm$ 0.010 \\
 &  & Moderate & 33.526 $\pm$ 1.429 & 0.900 $\pm$ 0.020 & 33.292 $\pm$ 0.603 & 0.913 $\pm$ 0.011 & 28.860 $\pm$ 0.785 & 0.871 $\pm$ 0.013 & \textbf{34.004 $\pm$ 1.288} & \textbf{0.933 $\pm$ 0.013} \\
 &  & Severe & 21.038 $\pm$ 0.529 & 0.505 $\pm$ 0.028 & 25.711 $\pm$ 0.667 & 0.721 $\pm$ 0.045 & 27.018 $\pm$ 0.693 & 0.821 $\pm$ 0.015 & \textbf{29.417 $\pm$ 0.985} & \textbf{0.881 $\pm$ 0.015}  \\
 % \cline{2-11}
 \cmidrule(lr){2-11}
 & \multirow{3}{*}{4.9} & Mild & \textbf{32.756 $\pm$ 0.514} & 0.915 $\pm$ 0.097 & 32.587 $\pm$ 0.661 & 0.911 $\pm$ 0.010 & 28.966 $\pm$ 0.803 & 0.877 $\pm$ 0.014 & 31.278 $\pm$ 0.908 & \textbf{0.918 $\pm$ 0.015} \\
 &  & Moderate & 30.496 $\pm$ 1.193 & 0.831 $\pm$ 0.022 & 31.066 $\pm$ 0.873 & 0.878 $\pm$ 0.080 & 27.156 $\pm$ 0.711 & 0.827 $\pm$ 0.016 & \textbf{31.093 $\pm$ 1.044} & \textbf{0.898 $\pm$ 0.030} \\
 &  & Severe & 21.420 $\pm$ 2.179 & 0.614 $\pm$ 0.072 & 22.404 $\pm$ 3.249 & 0.622 $\pm$ 0.088 & 25.905 $\pm$ 0.600 & 0.801 $\pm$ 0.017 & \textbf{28.121 $\pm$ 1.063} & \textbf{0.858 $\pm$ 0.068} \\
\bottomrule[1.5pt]
\end{tabular}
}
\label{tab:sim_results}
\end{table*}

Training data is derived from the FastMRI brain dataset~\cite{Knoll2020} and the Calgary-Campinas Brain MRI dataset (CC359)~\cite{Souza2018} as both datasets provide raw k-space data.
FastMRI provides a large number of volumes but only partial head coverage, whereas CC359 offers full-head MRI volumes but a limited sample size.
To leverage the complementary strengths of both datasets, we employ a two-stage training strategy: the diffusion model is first pre-trained on FastMRI to learn a generic 3D brain MRI prior, and subsequently fine-tuned on CC359 to adapt to full-head anatomy.
We use subject-independent splits for both FastMRI and CC359, following the official training and validation splits provided by each dataset. 
 For FastMRI, we restrict training to the T1-weighted volumes available in the dataset.

For both datasets, k-space data are normalized by the $99^{\text{th}}$ percentile of the root sum-of-squares (RSS) reconstruction magnitude.
Training is performed on complex-valued, coil-combined images, with coil sensitivities estimated using ESPIRiT~\cite{Uecker2013}.
Pre-training on FastMRI is conducted for 10,000 epochs using the Adam optimizer with an initial learning rate of $10^{-4}$, annealed to $10^{-6}$ via cosine scheduling.
Fine-tuning on CC359 is performed for an additional 2{,}000 epochs using the same optimization setup.

\section{Results}
\label{sec:results}

In this section, we evaluate the proposed MotionDPS method for accurate motion-trajectory estimation and reconstruction of 3D brain MRI volumes corrupted by motion and undersampling artifacts.
We first consider controlled motion simulation experiments using the FastMRI and CC359 datasets, and then demonstrate performance on the PMoC3D dataset~\cite{Wang2025}, which contains real motion-corrupted raw data.

To quantitatively evaluate reconstruction quality, we report the Structural Similarity Index (SSIM) and Peak Signal-to-Noise Ratio (PSNR)~\cite{Wang2025}. 
All experiments were performed on a single NVIDIA A100 GPU.

We compare MotionDPS against the following representative retrospective motion-correction methods:
\begin{itemize}
    \item \textit{Alternating optimization} (AltOpt)~\cite{cordero2016}: Iteratively alternates between wavelet-regularized image reconstruction and rigid motion parameter estimation.
    \item \textit{MotionTTT}~\cite{klug2024}: Estimates 3D motion parameters via test-time training by minimizing a data-consistency loss using a motion-free pretrained 2D U-Net (trained on CC359).
    \item \textit{E2E Stacked U-Nets} (StackedUnet)~\cite{almasni2021}: End-to-end slice-wise 2D U-Net architecture with neighboring-slice context to directly predict motion-corrected volumes (trained on CC359).
\end{itemize}

This benchmarking provides a direct comparison between the proposed model-based diffusion approach and established optimization- and learning-based motion correction methods.
Unlike AltOpt and MotionTTT, MotionDPS performs joint inference directly within the reverse-diffusion iterations and does not require additional outer optimization routines.

\subsection{Experiments with motion simulation}
\label{subsec:motion_sim}

We perform motion simulation experiments under a variety of k-space sampling trajectories and undersampling patterns to evaluate the robustness of MotionDPS across different acquisition settings. To simulate realistic motion, we sample smooth 6-DoF rigid trajectories from a Gaussian process prior with an RBF kernel~\cite{Wang2023}. 
The resulting trajectories~$\mathcal{T}$, corresponding to either individual k-space lines or batches of lines, enabling simulation of both intra- and inter-shot motion.

Performance is assessed with and without undersampling under three motion severity levels:
\begin{itemize}
    \item \textit{Mild}: translation of $\pm 3~\mathrm{mm}$ and rotation of $\pm 5^\circ$,
    \item \textit{Moderate}: translation of $\pm 6~\mathrm{mm}$ and rotation of $\pm 10^\circ$,
    \item \textit{Severe}: translation of $\pm 9~\mathrm{mm}$ and rotation of $\pm 15^\circ$.
\end{itemize}
We first generate synthetic inter-shot motion trajectories to corrupt k-space data from the fully sampled FastMRI dataset using 52 acquisition shots.
Reconstructions are compared against the original motion-free reference volumes. 
All experiments are conducted on the multicoil test split provided by the dataset, restricted to T1-weighted volumes.

To assess robustness under additional undersampling artifacts, we generate two 3D Cartesian undersampling masks with acceleration factors $R = 4$ and $R = 8$, each including 4\% auto-calibration lines (ACL), applied along the two phase-encoding dimensions. We adopt a linear circular sampling trajectory $\mathcal{T}$, in which the acquisition starts from the center of k-space and wraps around circularly.
We additionally evaluate centric, interleaved, and random orderings (each starting from the k-space center) and observe comparable reconstruction performance, indicating that MotionDPS is not sensitive to the specific k-space acquisition order.

\begin{figure}[t]
    \centering
    \includegraphics[width=1.0\linewidth]{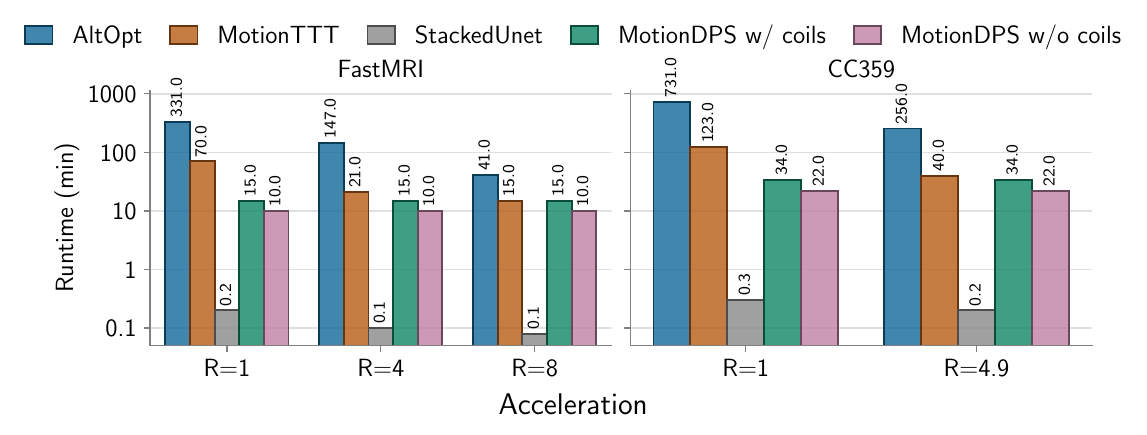}
    \caption{Runtimes of different reconstruction methods across acceleration factors. Two variants of MotionDPS are reported: with and without joint coil sensitivity estimation.}
    \label{fig:runtime_barplot}
\end{figure}

Fig.~\ref{fig:fastmri_coils} illustrates coil sensitivity estimation under different acceleration factors on the FastMRI dataset. 
MotionDPS maps closely match the motion-free ESPIRiT estimate. 
Under motion-corrupted k-space, ESPIRiT exhibits pronounced artifacts, whereas MotionDPS maintains smooth and artifact-free sensitivity estimates.

Finally, we generate synthetic inter-shot motion on the CC359 dataset using 52 acquisition shots. 
All experiments are conducted on the official validation split provided by the dataset.
We adopt the same motion severity levels described above and employ an interleaved k-space sampling trajectory $\mathcal{T}$, in which the central $3 \times 3$ region of k-space is acquired first. 
A 3D Cartesian Poisson-disc undersampling mask with an acceleration factor of $R = 4.9$ is applied to the two phase-encoding dimensions.
This sampling trajectory and undersampling mask are selected to replicate the acquisition geometry of PMoC3D raw data.

For all motion simulation experiments, we use the same set of hyperparameters. We run Algorithm~\ref{alg:motiondps} for $N=200$ reverse diffusion iterations with noise levels ranging from $\sigma_{\mathrm{min}}=0.002$ to $\sigma_{\mathrm{max}}=80.0$. 
We employ an exponentially decreasing schedule for $\zeta$, ranging from $1.0$ to $0.1$, to emphasize data consistency at early diffusion steps (high noise) and gradually reduce its influence at later steps to mitigate motion and undersampling artifacts. 
Coil regularization \(\gamma\) is set to \(200.0\) for FastMRI and \(2000.0\) for CC359.
We use different regularization weights for rotational (r) and translational (t) motion components to account for their different physical scales, with $\eta_{\mathrm{r}}=1000.0$ and $\eta_{\mathrm{t}}=50.0$. 
These hyperparameters are tuned using Optuna~\cite{Akiba2019} to balance reconstruction accuracy and computational cost.

To evaluate the baseline methods on FastMRI, we scale each k-space measurement using the median of the 99th percentile values of the RSS reconstructions computed from the CC359 dataset.
This normalization aligns the dynamic range of FastMRI with that of CC359, thereby enabling the reuse of most hyperparameters originally defined for CC359.

During the numerical experiments, we observed that the physics-based baselines exhibit high sensitivity to the acceleration factor \(R\).
To account for this behavior, the hyperparameters of AltOpt and MotionTTT were further optimized independently for each acceleration regime using Optuna~\cite{Akiba2019}. 
The resulting hyperparameter configurations are reported in Table~\ref{tab:ttt_altopt_hyper}, while the remaining hyperparameters were kept identical to those reported in Klug \etal~\cite{klug2024}.
In contrast, MotionDPS showed consistent performance across all acceleration factors; we therefore did not tune any hyperparameters.

\begin{table}[t]
    \caption{Hyperparameters used for MotionTTT and AltOpt across the motion-simulation experiments.}
    \label{tab:ttt_altopt_hyper}
    \centering
    \resizebox{\linewidth}{!}{
    \begin{tabular}{l|c|c|c}
    \toprule[1.5pt]
    Setting & MotionTTT LR & \(\ell_1\) LR & \(\ell_1\) Reg. \((\lambda\)) \\
    \midrule[1pt]
    FastMRI (\(R=1\)) & \(2.0\) & \(2.7\times10^{6}\) & \(1.6\times10^{-3}\)  \\
    FastMRI (\(R\in\{4,8\}\)) & \(4.0\) & \(1.1\times10^{7}\) & \(4\times 10^{-4}\) \\ 
    % FastMRI (\(R=8\)) & \(4.0\) & \(1.1\times10^{7}\) & \(4\times10^{-4}\) \\
    CC359 (\(R=1\)) & \(2.0\) & \(1.2\times10^{7}\) & \(4\times10^{-3}\) \\
    CC359 (\(R=4.9\)) & \(4.0\) & \(5\times10^7\) & \(10^{-3}\) \\
    \bottomrule[1.5pt]
    \end{tabular}}
\end{table}

% enabling the use of the default baseline hyperparameters without dataset-specific retuning. 
% Consequently, we run the baselines with the same hyperparameters reported in Klug \etal~\cite{klug2024}.

Table~\ref{tab:sim_results} list the mean and standard deviation for each metric for FastMRI and CC359. 
Reconstruction quality on FastMRI degrades gradually with increasing motion severity and undersampling acceleration, as expected. 
Despite this, MotionDPS maintains high PSNR/SSIM values even under severe motion and aggressive undersampling ($R=8$), indicating robust preservation of anatomical structures.
Results on CC359, which comprises more complex full-head anatomy and more realistic acquisition conditions, show lower performance reflecting the increased difficulty of the reconstruction task. 
Overall, MotionDPS maintains high reconstruction fidelity across regimes and remains accurate even under the most challenging setting, demonstrating strong robustness to the compounded effects of undersampling and motion.

Fig.~\ref{fig:runtime_barplot} reports the average reconstruction runtime for all evaluated methods across different acceleration factors. Since the compared baseline methods do not jointly estimate coil sensitivity maps during reconstruction, we include two variants of MotionDPS: (i) joint coil sensitivity estimation, and (ii) reconstruction using ESPIRiT coil sensitivity maps computed from motion-free reference k-space data.
As expected, the feed-forward StackedUnet exhibits the lowest computational cost due to its single-pass inference formulation. 
Among the iterative physics-based reconstruction methods, MotionDPS achieves the lowest runtime, even with joint coil sensitivity estimation. Furthermore, the computational cost of the baseline methods increases at lower acceleration factors due to the larger amount of acquired k-space measurements involved in the data-consistency updates. 
In contrast, the runtime of MotionDPS remains constant across acceleration factors.

% Interestingly, AltOpt and MotionTTT show weaker performance in the fully sampled setting ($R=1$) than in undersampled regimes.
% We hypothesize that the hyperparameters of these methods were not specifically adapted for the fully sampled case. 
% In our experiments, we retained the default hyperparameters provided by the original authors and did not perform additional tuning for different acceleration factors.
% In contrast, MotionDPS does not exhibit this sensitivity to the acceleration factor, as the same set of hyperparameters is used across all sampling regimes without additional tuning.

Fig.~\ref{fig:cc359_recon_sub8} shows representative results on a CC359 scan under severe motion.
The RSS reconstruction exhibits pronounced blurring and ghosting artifacts that obscure fine anatomical details. 
MotionTTT fails to adequately compensate for motion, resulting in residual artifacts, structural distortions, and inaccurate motion estimates. In contrast, MotionDPS restores the global brain structure and sharp tissue boundaries, closely matching the motion-free reference. 
Remaining reconstruction errors are primarily confined to high-contrast edges and peripheral regions. 
The recovered motion trajectories closely follow the ground-truth signals, with only slight deviations in the translational components, whereas MotionTTT exhibit substantial deviations from the ground truth.

\begin{table}[t]
\caption{PMoC3D real-motion results: PSNR (dB) and SSIM for mild, moderate, and severe motion; best scores are highlighted in bold.}
\centering
\renewcommand{\arraystretch}{1.5}
\resizebox{\linewidth}{!}{
\begin{tabular}{c|l|cc|cc|cc|cc}
\toprule[1.5pt]
\multirow{2}{*}{} & \multirow{2}{*}{Scan ID} & \multicolumn{2}{c|}{AltOpt} & \multicolumn{2}{c|}{MotionTTT} & \multicolumn{2}{c|}{StackedUnet} & \multicolumn{2}{c}{MotionDPS{\footnotesize (our)}} \\
&  & PSNR & SSIM & PSNR & SSIM & PSNR & SSIM & PSNR & SSIM \\
\midrule[1pt]
\multirow{8}{*}{\rotatebox{90}{Mild}}
& S1\_2 & 30.077 & 0.954 & 33.333 & 0.956 & 33.544 & 0.953 & \textbf{33.781} & \textbf{0.958} \\
& S7\_3 & 32.085 & 0.960 & 32.117 & 0.960 & 31.422 & 0.957 & \textbf{32.572} & \textbf{0.961} \\
& S7\_1 & 32.276 & 0.960 & 32.512 & 0.961 & 31.990 & 0.956 & \textbf{33.104} & \textbf{0.962} \\
& S3\_3 & 31.283 & 0.955 & 31.239 & 0.956 & 30.869 & 0.952 & \textbf{31.950} & \textbf{0.957} \\
& S1\_3 & 32.253 & 0.947 & 33.104 & \textbf{0.955} & 32.985 & 0.948 & \textbf{33.376} & \textbf{0.955} \\
& S4\_2 & 31.095 & 0.954 & 31.212 & 0.955 & 29.249 & 0.946 & \textbf{32.156} & \textbf{0.956} \\
& S5\_2 & 32.457 & 0.960 & 32.585 & 0.960 & 31.732 & 0.956 & \textbf{32.971} & \textbf{0.961} \\
& S2\_3 & 31.233 & 0.958 & 31.488 & 0.960 & 31.033 & 0.950 & \textbf{32.315} & \textbf{0.961} \\
\midrule[1pt]
\multirow{8}{*}{\rotatebox{90}{Moderate}}
& S3\_2 & 30.844 & 0.952 & 30.990 & 0.953 & 29.566 & 0.940 & \textbf{31.628} & \textbf{0.955} \\
& S4\_1 & 30.244 & 0.951 & 30.473 & 0.953 & 29.078 & 0.942 & \textbf{32.422} & \textbf{0.956} \\
& S7\_2 & 30.550 & 0.954 & 30.909 & 0.957 & 28.430 & 0.934 & \textbf{31.615} & \textbf{0.958} \\
& S5\_3 & 31.555 & 0.956 & 32.189 & \textbf{0.961} & 29.762 & 0.934 & \textbf{32.729} & \textbf{0.961} \\
& S6\_2 & 31.146 & 0.958 & 31.909 & \textbf{0.962} & 29.740 & 0.941 & \textbf{32.071} & 0.961 \\
& S2\_2 & 31.325 & 0.959 & 31.745 & \textbf{0.961} & 30.053 & 0.940 & \textbf{32.256} & \textbf{0.961} \\
& S8\_2 & 29.933 & 0.954 & 30.510 & 0.959 & 27.290 & 0.937 & \textbf{31.672} & \textbf{0.961} \\
& S6\_1 & 30.149 & 0.954 & 30.403 & 0.958 & 28.957 & 0.941 & \textbf{31.881} & \textbf{0.960} \\
\midrule[1pt]
\multirow{8}{*}{\rotatebox{90}{Severe}}
& S4\_3 & 28.570 & 0.946 & 29.088 & 0.949 & 26.458 & 0.936 & \textbf{30.504} & \textbf{0.952} \\
& S3\_1 & 30.403 & 0.951 & 30.638 & 0.953 & 28.105 & 0.928 & \textbf{31.384} & \textbf{0.954} \\
& S5\_1 & 31.772 & 0.958 & 32.143 & \textbf{0.960} & 29.984 & 0.942 & \textbf{32.675} & \textbf{0.960} \\
& S1\_1 & 29.917 & 0.926 & 31.536 & \textbf{0.949} & 29.960 & 0.916 & \textbf{32.226} & \textbf{0.949} \\
& S8\_1 & 30.041 & 0.955 & 30.478 & \textbf{0.959} & 25.643 & 0.922 & \textbf{30.993} & 0.957 \\
& S6\_3 & 29.517 & 0.950 & 30.098 & \textbf{0.954} & 27.120 & 0.931 & \textbf{30.751} & \textbf{0.954} \\
& S8\_3 & 30.835 & 0.958 & 31.031 & \textbf{0.960} & 26.384 & 0.927 & \textbf{31.393} & \textbf{0.960} \\
& S2\_1 & 27.838 & 0.939 & 27.460 & 0.944 & 25.430 & 0.902 & \textbf{30.286} & \textbf{0.951} \\
\midrule[1pt]
& Average & 30.850 & 0.953 & 31.216 & 0.957 & 29.366 & 0.939 & \textbf{32.030} & \textbf{0.958} \\
\bottomrule[1.5pt]
\end{tabular}
}
\label{tab:pmoc3d_results}
\end{table}

\begin{figure*}[t]
\centering
\resizebox{0.93\linewidth}{!}{
\begin{tikzpicture}[
  font=\small,
  panel/.style={inner sep=0pt, outer sep=0pt}
]

% ---- layout parameters ----
\def\W{2.3cm}
\def\H{2.3cm}
\def\xgap{0.03cm}
\def\ygap{0.03cm}

\def\blockgap{0.6cm} 
\def\ySTwo{0cm}
\def\ySFour{-\blockgap}
\def\ySSeven{-2*\blockgap}

% ---------------- Column headers ----------------
\node at ({0*(\W+\xgap)+0.5*\W}, {9.1*(\H+\ygap)}) {Reference};
\node at ({1*(\W+\xgap)+0.5*\W}, {9.1*(\H+\ygap)}) {RSS};
\node at ({2*(\W+\xgap)+0.5*\W}, {9.1*(\H+\ygap)}) {AltOpt};
\node at ({3*(\W+\xgap)+0.5*\W}, {9.1*(\H+\ygap)}) {MotionTTT};
\node at ({4*(\W+\xgap)+0.5*\W}, {9.1*(\H+\ygap)}) {StackedUnet};
\node at ({5*(\W+\xgap)+0.5*\W}, {9.1*(\H+\ygap)}) {MotionDPS};

% ---------------- Row labels ----------------
\node[rotate=90] at (-0.3cm, {8*(\H+\ygap)+0.5*\H + \ySTwo}) {Coronal};
\node[rotate=90] at (-0.3cm, {7*(\H+\ygap)+0.5*\H + \ySTwo}) {Error};
\node[rotate=90] at (-0.3cm, {6.5*(\H+\ygap) + \ySTwo}) {Motion};

\node[rotate=90] at (-0.3cm, {5*(\H+\ygap)+0.5*\H + \ySFour}) {Sagittal};
\node[rotate=90] at (-0.3cm, {4*(\H+\ygap)+0.5*\H + \ySFour}) {Error};
\node[rotate=90] at (-0.3cm, {3.5*(\H+\ygap) + \ySFour}) {Motion};

% ---------------- Scan ID labels ----------------
\node[rotate=90] at (-0.8cm, {7.5*(\H+\ygap) + \ySTwo}) {Severe (S2\_1)};
\node[rotate=90] at (-0.8cm, {4.5*(\H+\ygap) + \ySFour}) {Moderate (S4\_1)};
% \node[rotate=90] at (-0.8cm, {1.5*(\H+\ygap) + \ySSeven}) {Mild (S7\_1)};

% ---------------- Helpers ----------------
\newcommand{\panelimg}[4]{% col, row, yshift, filename
  \node[panel, anchor=south west]
    at ({#1*(\W+\xgap)},{#2*(\H+\ygap) + #3}) {%
      \includegraphics[width=\W,height=\H]{#4}%
    };
}

\newcommand{\panelimgpsnr}[5]{% col, row, yshift, filename, psnr
  \node[panel, anchor=south west] (img-#1-#2)
    at ({#1*(\W+\xgap)},{#2*(\H+\ygap) + #3}) {%
      \includegraphics[width=\W,height=\H]{#4}%
    };
  \node[anchor=north east, xshift=-1.5pt, yshift=-1.5pt,
        fill=black, rounded corners=1pt, inner sep=1.5pt]
    at (img-#1-#2.north east)
    {\textcolor{green!100!black}{\bfseries\footnotesize #5}};
}

% motion row spanning all columns
\newcommand{\panelmotion}[3]{% row, yshift, filename
  \node[panel, anchor=south]
    at ({0.5*6*\W + 0.5*5*\xgap}, {#1*(\H+\ygap) + #2}) {%
      \includegraphics[width=0.75\linewidth]{#3}%
    };
}

% ==================== S2_1 (Coronal) ====================
\panelimg{0}{8}{\ySTwo}{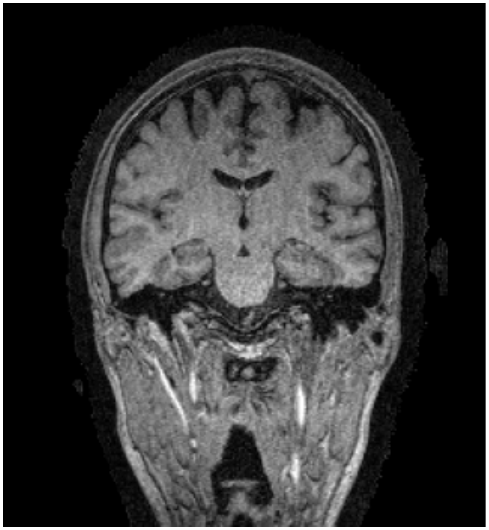}
\panelimg{1}{8}{\ySTwo}{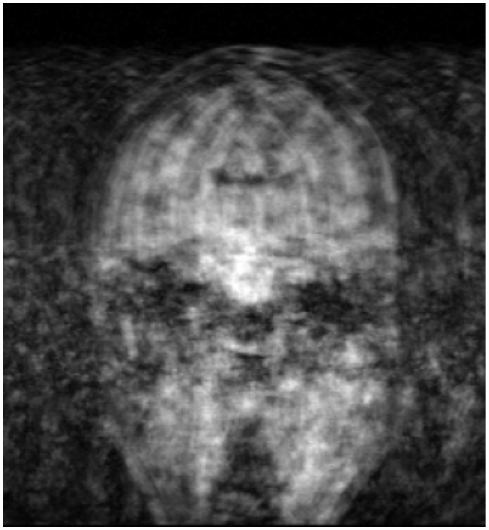}
\panelimgpsnr{2}{8}{\ySTwo}{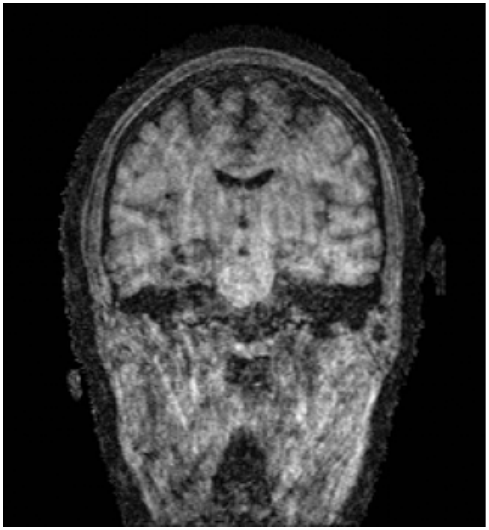}{27.838}
\panelimgpsnr{3}{8}{\ySTwo}{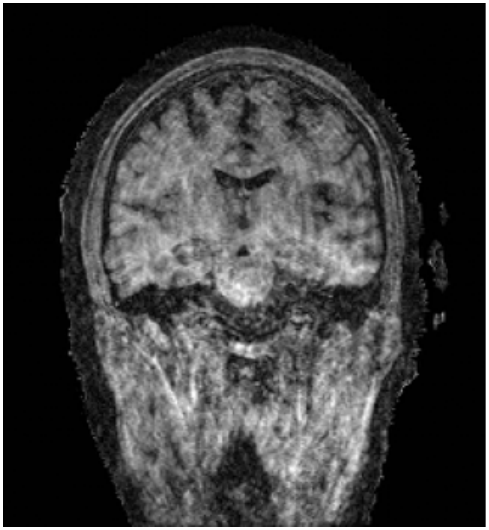}{27.460}
\panelimgpsnr{4}{8}{\ySTwo}{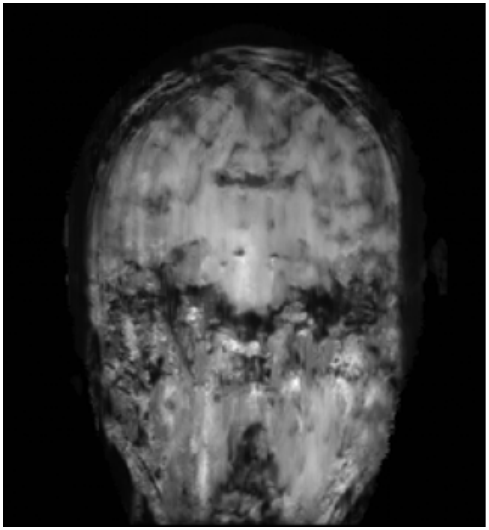}{25.430}
\panelimgpsnr{5}{8}{\ySTwo}{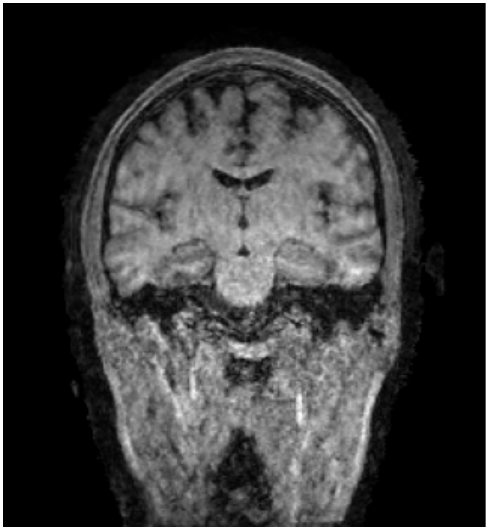}{30.286}

\node[panel, anchor=south west] at (0,{7*(\H+\ygap) + \ySTwo}) {
  \makebox[\W][c]{%
    \includegraphics[width=0.12\linewidth]{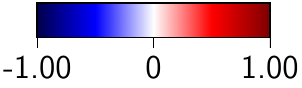}}
};
\panelimg{1}{7}{\ySTwo}{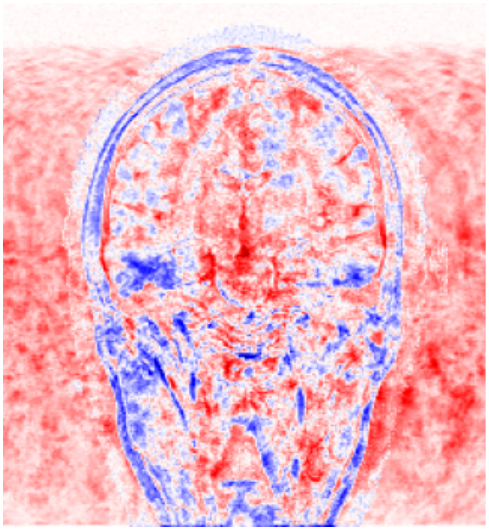}
\panelimg{2}{7}{\ySTwo}{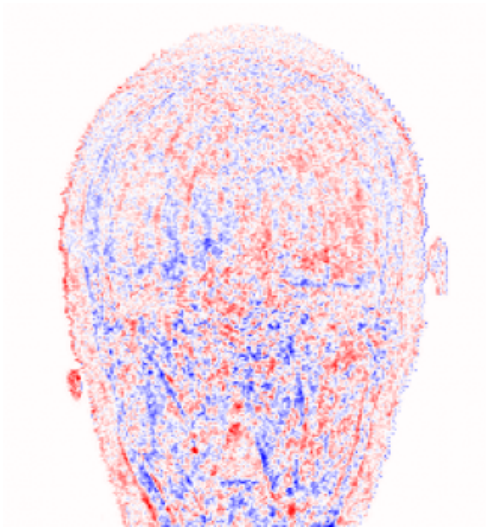}
\panelimg{3}{7}{\ySTwo}{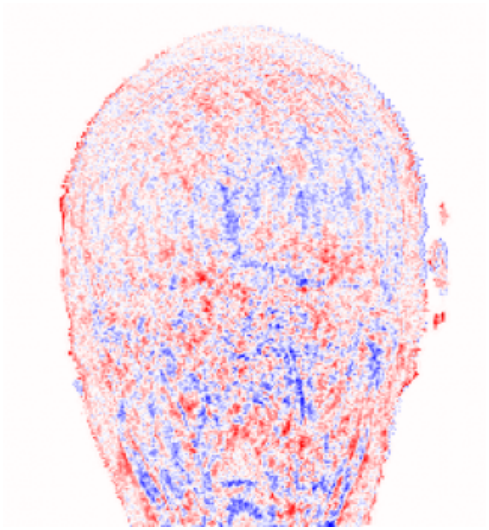}
\panelimg{4}{7}{\ySTwo}{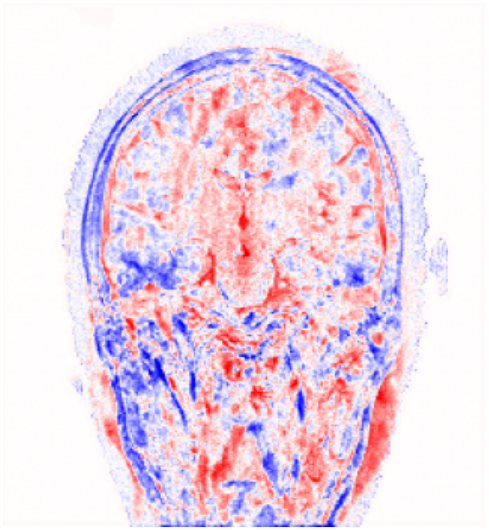}
\panelimg{5}{7}{\ySTwo}{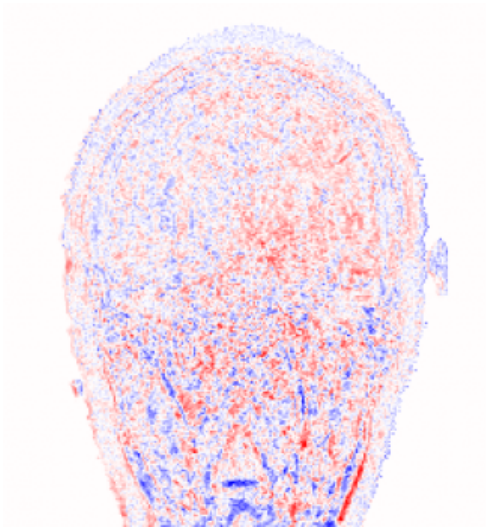}

\panelmotion{5.75}{\ySTwo}{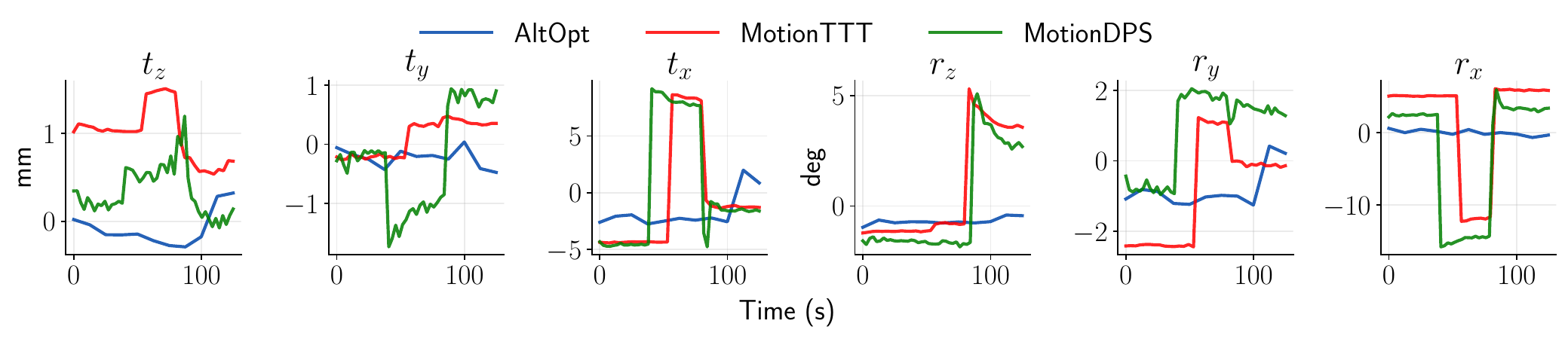}

% ==================== S4_1 (Sagittal) ====================
\panelimg{0}{5}{\ySFour}{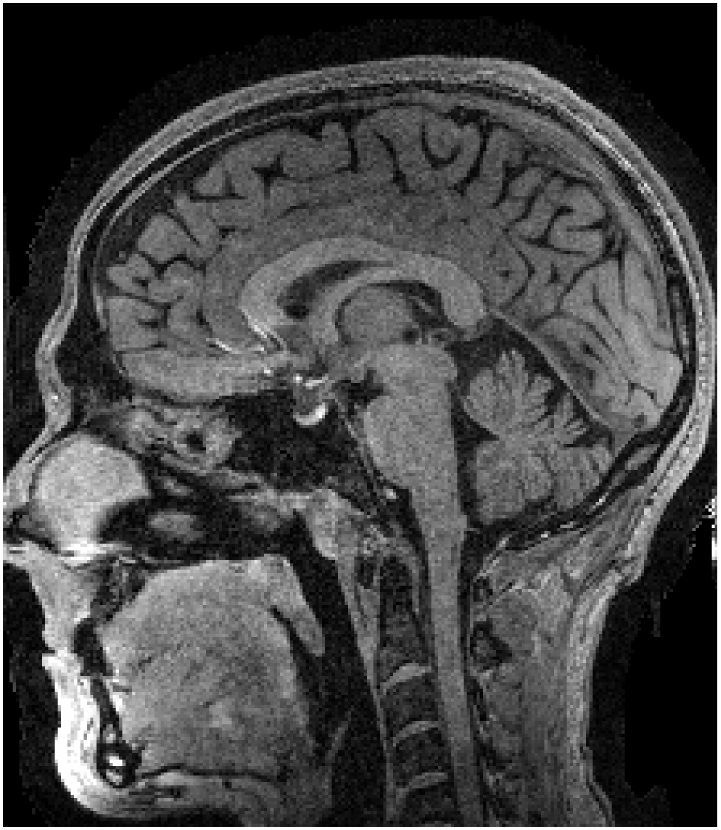}
\panelimg{1}{5}{\ySFour}{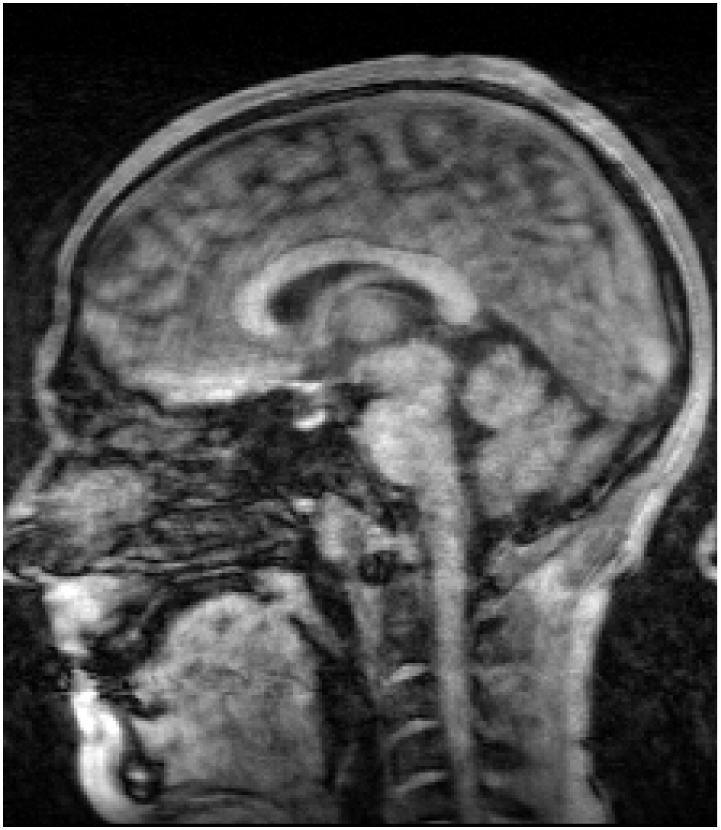}
\panelimgpsnr{2}{5}{\ySFour}{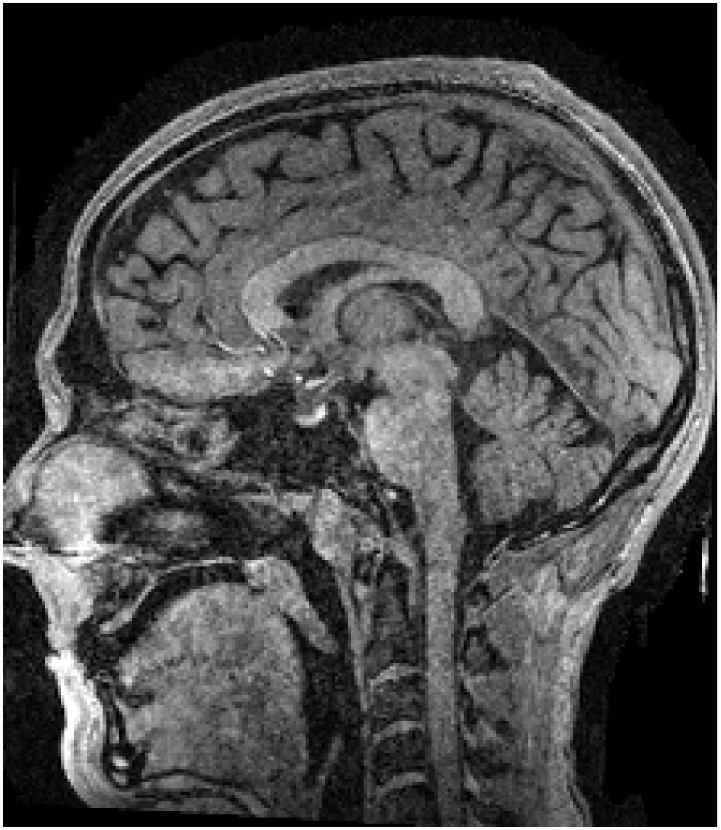}{30.244}
\panelimgpsnr{3}{5}{\ySFour}{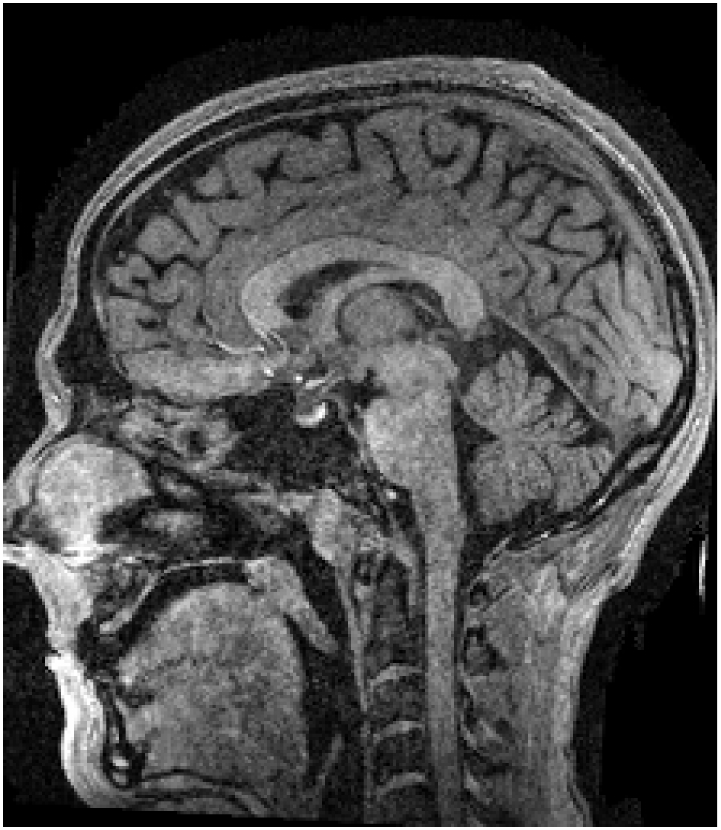}{30.473}
\panelimgpsnr{4}{5}{\ySFour}{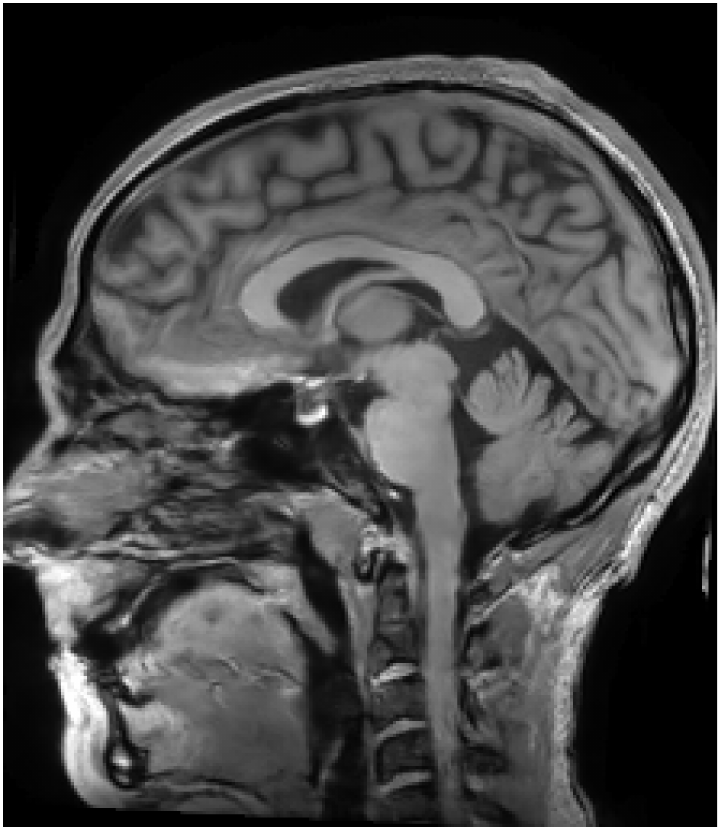}{29.078}
\panelimgpsnr{5}{5}{\ySFour}{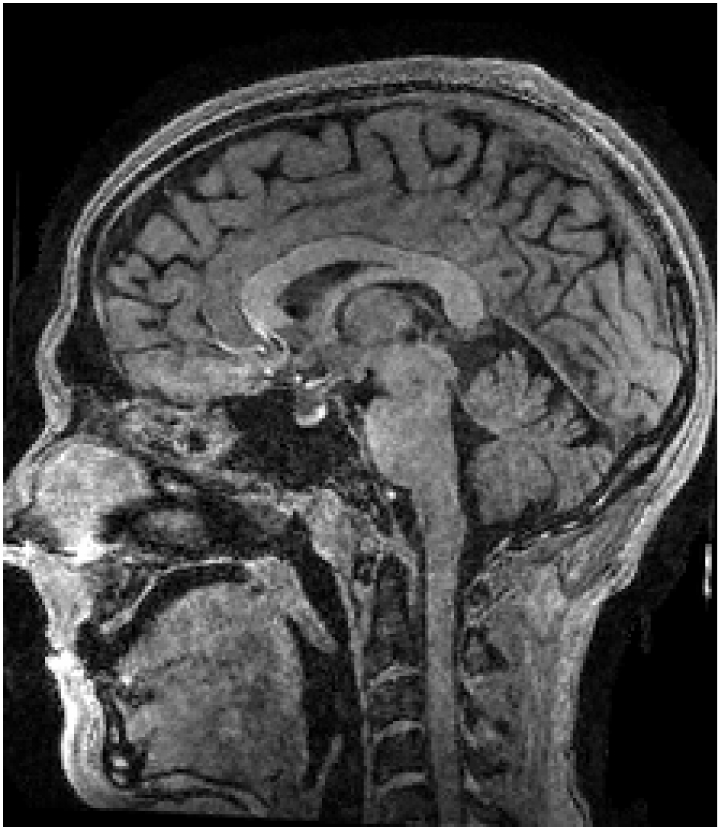}{32.422}

\node[panel, anchor=south west] at (0,{4*(\H+\ygap) + \ySFour}) {
  \makebox[\W][c]{%
    \includegraphics[width=0.12\linewidth]{images/pmoc3d_S2_1/error_colorbar.pdf}}
};
\panelimg{1}{4}{\ySFour}{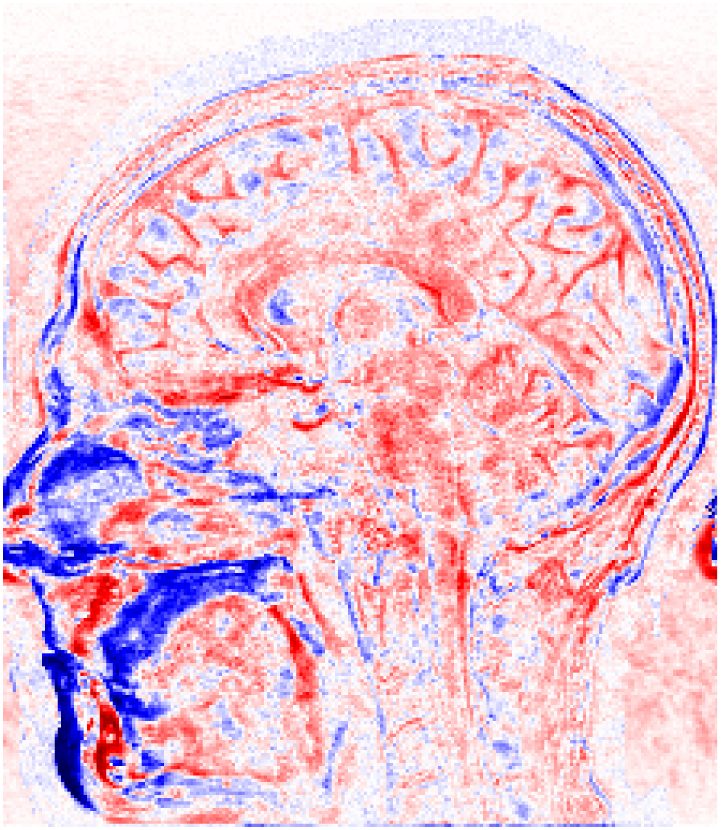}
\panelimg{2}{4}{\ySFour}{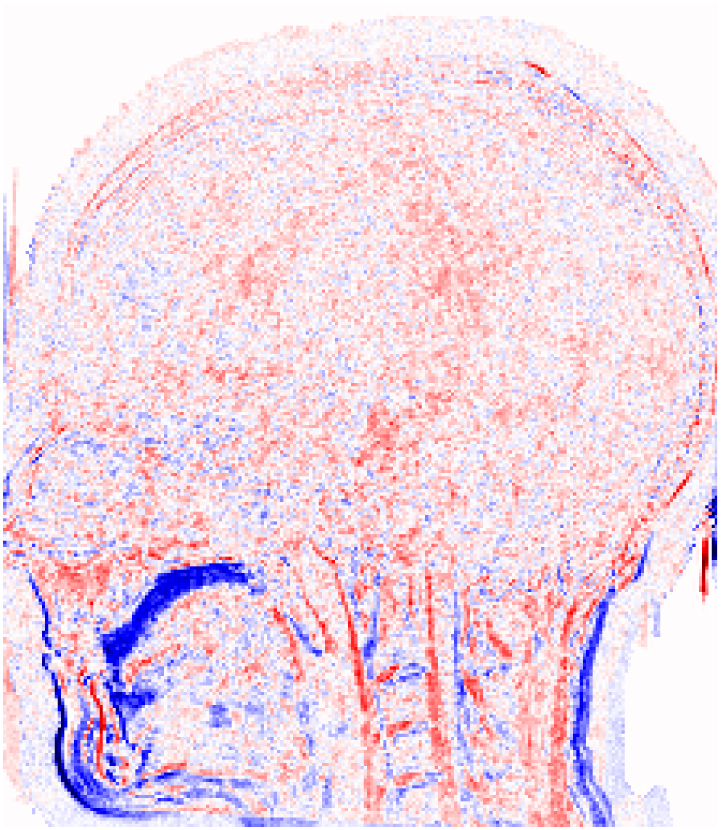}
\panelimg{3}{4}{\ySFour}{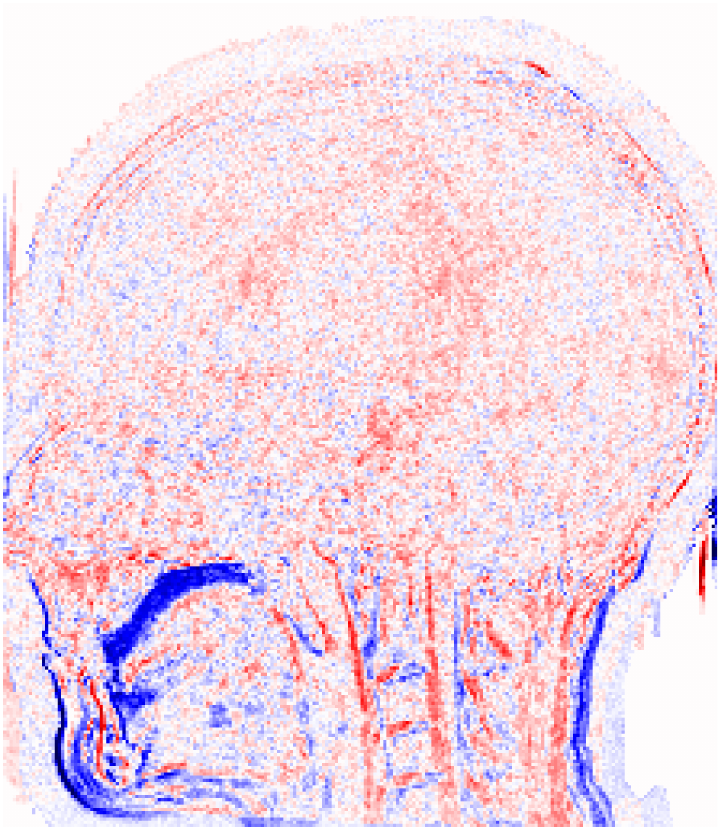}
\panelimg{4}{4}{\ySFour}{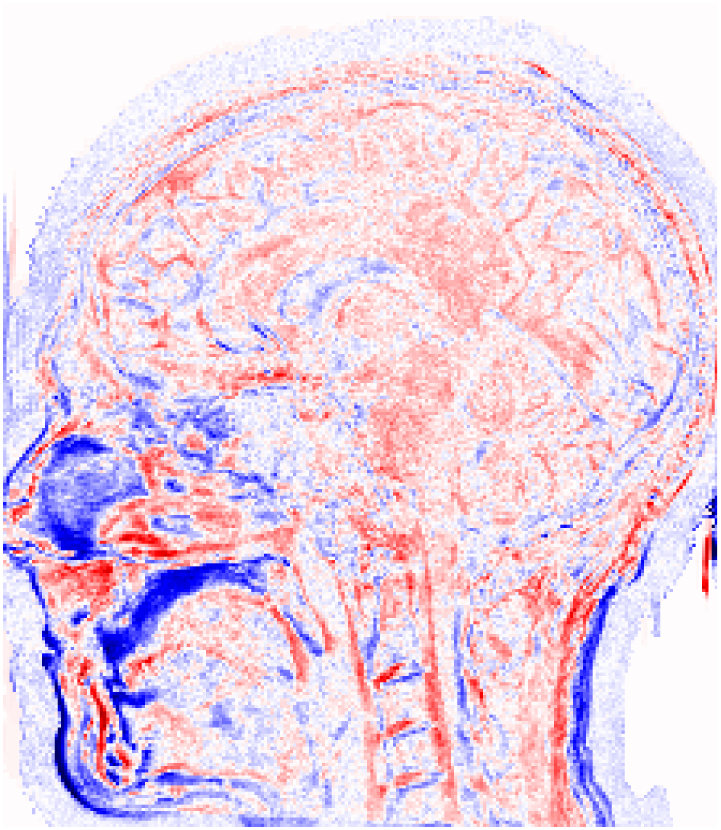}
\panelimg{5}{4}{\ySFour}{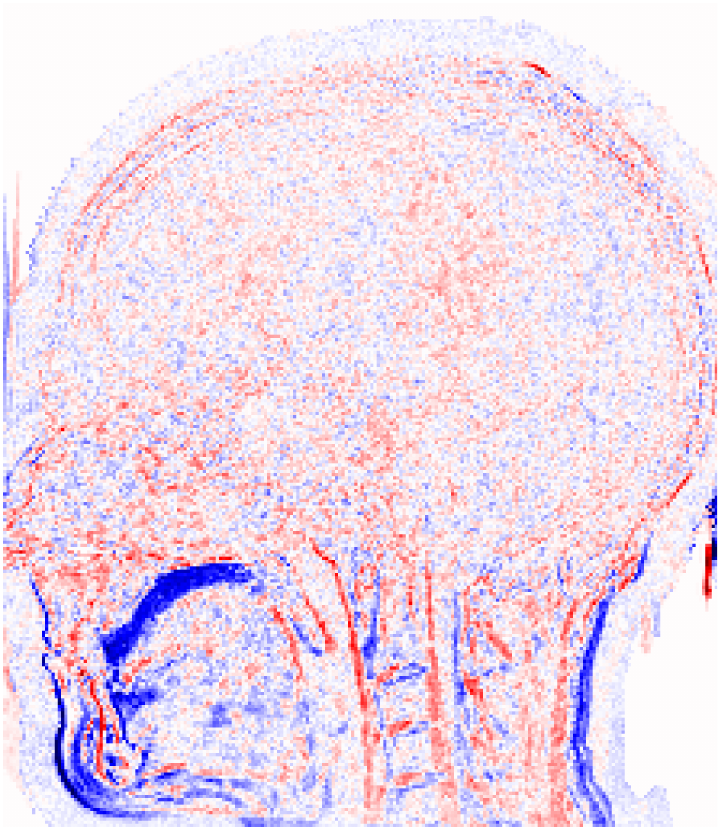}

\panelmotion{2.8}{\ySFour}{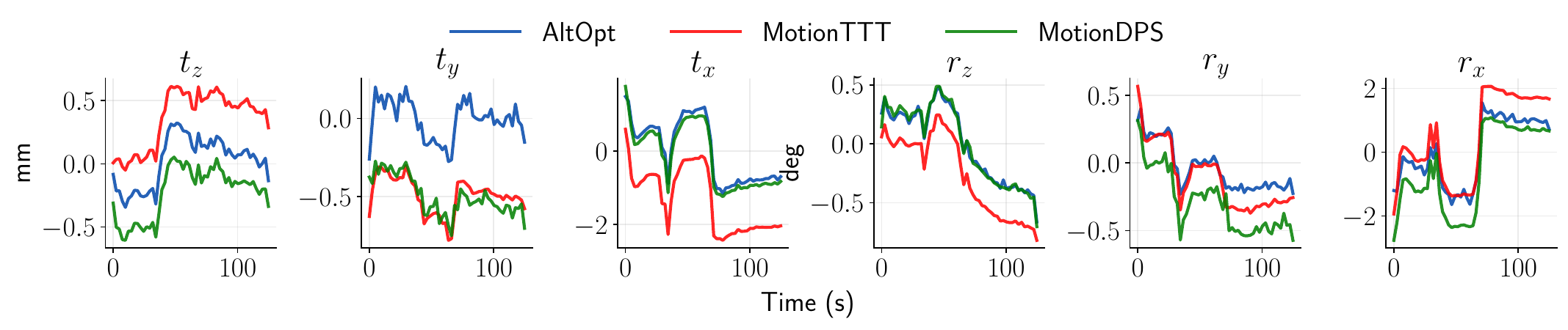}

\end{tikzpicture}
}
\caption{Qualitative reconstruction results for two scans of the PMoC3D dataset. For each scan, the reconstruction, error map, and estimated motion parameters are shown. PSNR scores (green, top-right) are computed w.r.t.~the motion-free reference.}
\label{fig:pmoc3d_recon}
\end{figure*}

\begin{figure*}[t]
\centering
\resizebox{0.93\linewidth}{!}{
\begin{tikzpicture}[
  font=\small,
  panel/.style={inner sep=0pt, outer sep=0pt}
]

% ---- layout parameters ----
\def\W{2.3cm}
\def\H{2.3cm}
\def\xgap{0.03cm}
\def\ygap{0.03cm}

\def\blockgap{0.6cm} 
\def\ySTwo{0cm}
\def\ySFour{-\blockgap}
\def\ySSeven{-2*\blockgap}

% ---------------- Column headers ----------------
\node at ({0*(\W+\xgap)+0.5*\W}, {9.1*(\H+\ygap)}) {Reference};
\node at ({1*(\W+\xgap)+0.5*\W}, {9.1*(\H+\ygap)}) {RSS};
\node at ({2*(\W+\xgap)+0.5*\W}, {9.1*(\H+\ygap)}) {AltOpt};
\node at ({3*(\W+\xgap)+0.5*\W}, {9.1*(\H+\ygap)}) {MotionTTT};
\node at ({4*(\W+\xgap)+0.5*\W}, {9.1*(\H+\ygap)}) {StackedUnet};
\node at ({5*(\W+\xgap)+0.5*\W}, {9.1*(\H+\ygap)}) {MotionDPS};

% ---------------- Row labels ----------------
\node[rotate=90] at (-0.3cm, {8*(\H+\ygap)+0.5*\H + \ySTwo}) {Axial};
\node[rotate=90] at (-0.3cm, {7*(\H+\ygap)+0.5*\H + \ySTwo}) {Error};
\node[rotate=90] at (-0.3cm, {6.5*(\H+\ygap) + \ySTwo}) {Motion};

\node[rotate=90] at (-0.3cm, {5*(\H+\ygap)+0.5*\H + \ySFour}) {Sagittal};
\node[rotate=90] at (-0.3cm, {4*(\H+\ygap)+0.5*\H + \ySFour}) {Error};
\node[rotate=90] at (-0.3cm, {3.5*(\H+\ygap) + \ySFour}) {Motion};

% ---------------- Scan ID labels ----------------
\node[rotate=90] at (-0.8cm, {7.5*(\H+\ygap) + \ySTwo}) {Severe (S8\_1)};
\node[rotate=90] at (-0.8cm, {4.5*(\H+\ygap) + \ySFour}) {Moderate (S6\_2)};
% \node[rotate=90] at (-0.8cm, {1.5*(\H+\ygap) + \ySSeven}) {Mild (S7\_1)};

% ---------------- Helpers ----------------
\newcommand{\panelimg}[4]{% col, row, yshift, filename
  \node[panel, anchor=south west]
    at ({#1*(\W+\xgap)},{#2*(\H+\ygap) + #3}) {%
      \includegraphics[width=\W,height=\H]{#4}%
    };
}

\newcommand{\panelimgpsnr}[5]{% col, row, yshift, filename, psnr
  \node[panel, anchor=south west] (img-#1-#2)
    at ({#1*(\W+\xgap)},{#2*(\H+\ygap) + #3}) {%
      \includegraphics[width=\W,height=\H]{#4}%
    };
  \node[anchor=north east, xshift=-1.5pt, yshift=-1.5pt,
        fill=black, rounded corners=1pt, inner sep=1.5pt]
    at (img-#1-#2.north east)
    {\textcolor{green!100!black}{\bfseries\footnotesize #5}};
}

% motion row spanning all columns
\newcommand{\panelmotion}[3]{% row, yshift, filename
  \node[panel, anchor=south]
    at ({0.5*6*\W + 0.5*5*\xgap}, {#1*(\H+\ygap) + #2}) {%
      \includegraphics[width=0.75\linewidth]{#3}%
    };
}

\panelimg{0}{8}{\ySTwo}{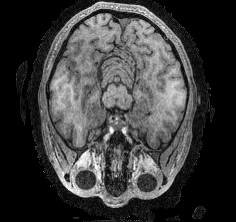}
\panelimg{1}{8}{\ySTwo}{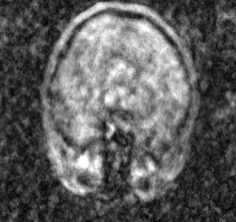}
\panelimgpsnr{2}{8}{\ySTwo}{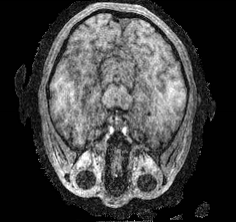}{0.955}
\panelimgpsnr{3}{8}{\ySTwo}{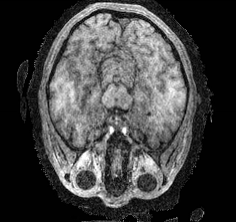}{0.959}
\panelimgpsnr{4}{8}{\ySTwo}{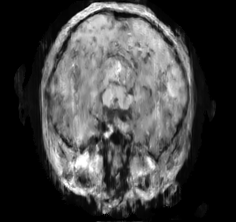}{0.922}
\panelimgpsnr{5}{8}{\ySTwo}{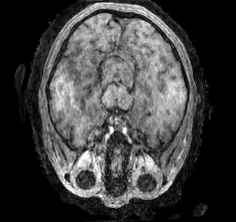}{0.957}

\node[panel, anchor=south west] at (0,{7*(\H+\ygap) + \ySTwo}) {
  \makebox[\W][c]{%
    \includegraphics[width=0.12\linewidth]{images/pmoc3d_S2_1/error_colorbar.pdf}}
};
\panelimg{1}{7}{\ySTwo}{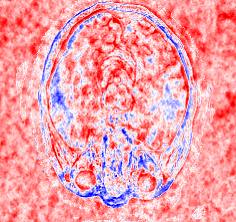}
\panelimg{2}{7}{\ySTwo}{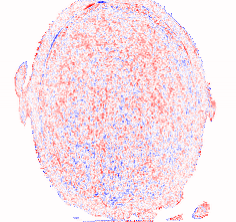}
\panelimg{3}{7}{\ySTwo}{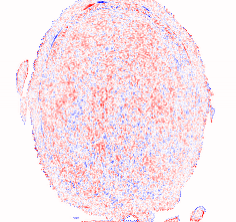}
\panelimg{4}{7}{\ySTwo}{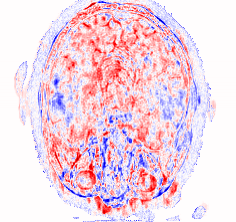}
\panelimg{5}{7}{\ySTwo}{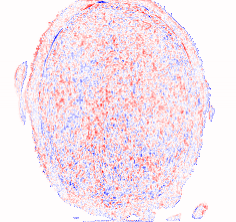}

\panelmotion{5.75}{\ySTwo}{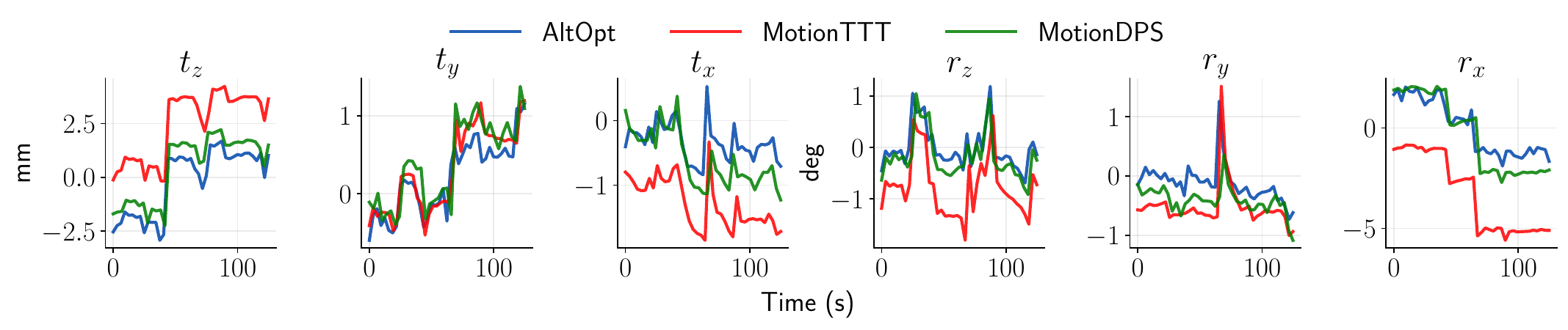}

\panelimg{0}{5}{\ySFour}{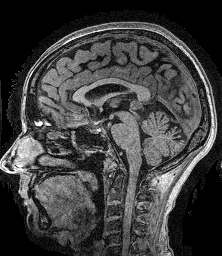}
\panelimg{1}{5}{\ySFour}{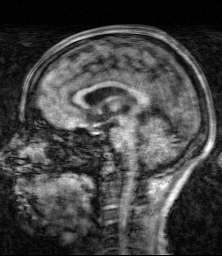}
\panelimgpsnr{2}{5}{\ySFour}{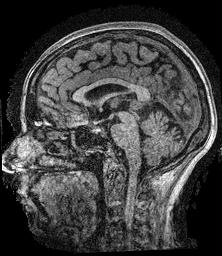}{0.958}
\panelimgpsnr{3}{5}{\ySFour}{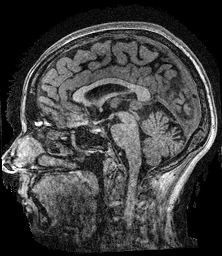}{0.962}
\panelimgpsnr{4}{5}{\ySFour}{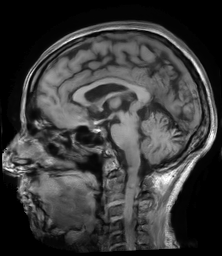}{0.941}
\panelimgpsnr{5}{5}{\ySFour}{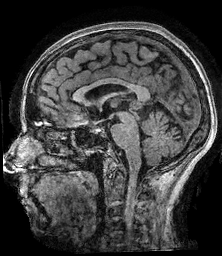}{0.961}

\node[panel, anchor=south west] at (0,{4.05*(\H+\ygap) + \ySFour}) {
  \makebox[\W][c]{%
    \includegraphics[width=0.12\linewidth]{images/pmoc3d_S2_1/error_colorbar.pdf}}
};
\panelimg{1}{4}{\ySFour}{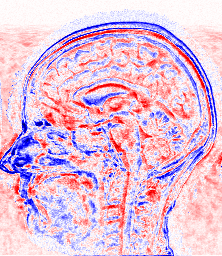}
\panelimg{2}{4}{\ySFour}{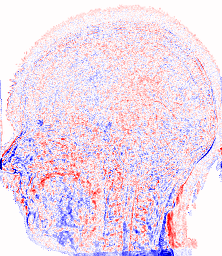}
\panelimg{3}{4}{\ySFour}{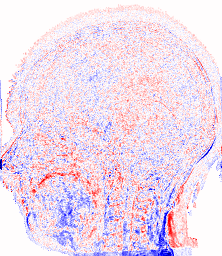}
\panelimg{4}{4}{\ySFour}{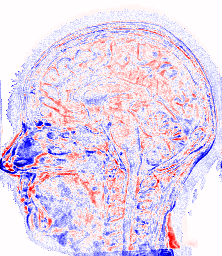}
\panelimg{5}{4}{\ySFour}{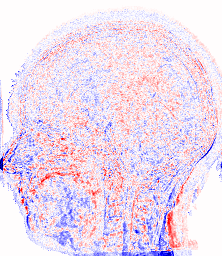}

\panelmotion{2.8}{\ySFour}{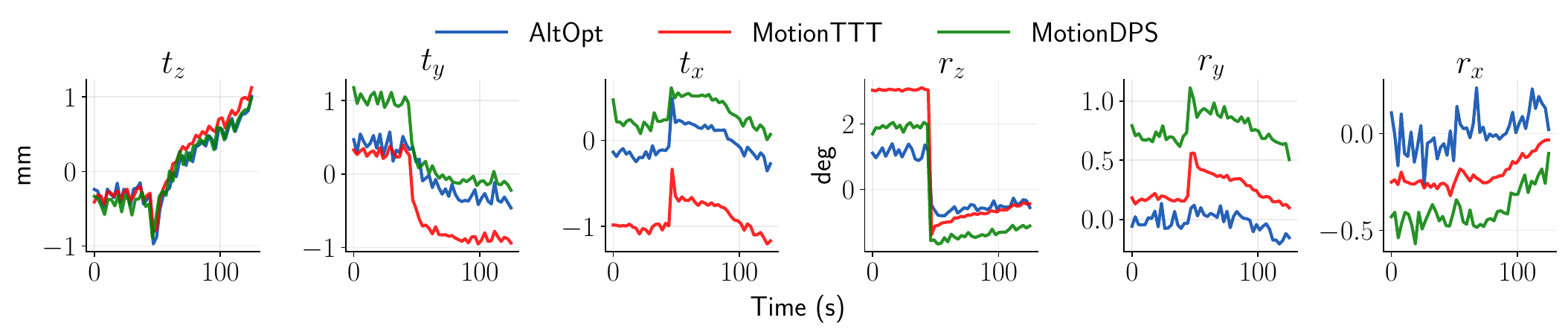}

\end{tikzpicture}
}
\caption{Qualitative analysis of challenging cases from the PMoC3D dataset where MotionDPS does not outperform all baselines in terms of SSIM. The top-right green values indicate the corresponding SSIM scores.}
\label{fig:pmoc3d_recon_lower_ssim}
\end{figure*}

\subsection{Experiments with real motion}
We conduct real-motion experiments using the PMoC3D dataset~\cite{Wang2025}, which comprises 27 volumes from 8 healthy subjects.
The dataset is used exclusively for evaluation, with no samples used during training of the diffusion prior. 
The data were acquired on a Philips Ingenia Elition 3T scanner, providing 3D T1-weighted volumes of size $k_z\times k_y\times k_x= 512\times 236\times 222$ with $C=13$ receiver coils at an isotropic voxel size of $1~\mathrm{mm}$. 
Acquisition was performed with an undersampling factor of $R=4.94$ along the two phase-encoding directions $k_y$ and $k_x$, over 52 shots following a quasi-random k-space sampling trajectory $\mathcal{T}$ in which the central $3\times 3$ region of k-space is acquired first. 
Following Klug \etal~\cite{klug2024}, we crop the data along the readout dimension $k_z$ to a size of 256 by subsampling every second voxel to reduce computational cost, and normalize each k-space volume by the $99^{\mathrm{th}}$ percentile of the RSS reconstruction.

The dataset provides four scans per subject: one motion-free and three motion-corrupted scans. Scans are labeled as $\mathrm{S}\{\mathrm{subject}\}\_\{\mathrm{scan}\}$ where $\mathrm{subject}\in \{1,...,8\}$ and $\mathrm{scan}\in \{0,...,3\}$, with $\mathrm{scan}=0$ denoting the motion-free reference. Details on the generation of the motion-corrupted scans can be found in Wang~\etal~\cite{Wang2025}.
For quantitative evaluation, we adopt the same motion severity categorization (\emph{mild}, \emph{moderate}, and \emph{severe}) as in Wang \etal~\cite{Wang2025} and use the $\ell_1$-regularized reconstruction of the motion-free scan as the reference. 

We apply MotionDPS and each baseline to all motion-corrupted scans and assess reconstruction quality with respect to the $\ell_1$-reconstruction of the motion-free reference.
We post-process all reconstructed volumes prior to evaluation: (i) rigidly register each reconstruction to the reference volume; (ii) extract a brain mask from the reference and apply it to both volumes; and (iii) perform intensity normalization by scaling each volume using its $99.9^{\mathrm{th}}$ percentile.

For all experiments, we use the same hyperparameter configuration as in Section~\ref{subsec:motion_sim}, with the exception of the number of reverse diffusion iterations $N$.
For \emph{severe} motion scans we keep $N=200$, while for all remaining scans we use $N=100$.
% As shown in Fig.~\ref{fig:convergence}, PSNR values for \emph{mild} and \emph{moderate} motion peak at approximately $N=100$, whereas reconstructions under \emph{severe} motion continue to improve with additional iterations.
Additionally, following Klug~\etal~\cite{klug2024}, motion states with a data consistency (DC) value greater than 0.75 are removed from the reconstruction process in the final $40$ iterations.

Table~\ref{tab:pmoc3d_results} reports quantitative reconstruction results on PMoC3D, grouped by \emph{mild}, \emph{moderate}, and \emph{severe} motion regimes.
MotionDPS consistently yields higher PSNR/SSIM values, indicating improved structural fidelity and better preservation of anatomical details. 
To assess whether these improvements are statistically significant, we perform paired Wilcoxon signed-rank tests comparing MotionDPS against each baseline for PSNR and SSIM, resulting in six hypothesis tests (3 baselines \(\times\) 2 metrics). 
$p$-values are corrected using the Benjamini--Hochberg procedure. 
Across all 24 motion-corrupted scans, MotionDPS significantly outperforms all baselines after correction. 
Mean PSNR gains relative to AltOpt, MotionTTT, and StackedUnet were \(1.305~\mathrm{dB}\), \(0.813~\mathrm{dB}\), and \(2.664~\mathrm{dB}\), respectively, with all comparisons remaining significant after correction. 
Corresponding SSIM improvements were \(0.0047\), \(0.0011\), and \(0.0187\), respectively, and all remained significant after correction, with the least significant comparison observed against MotionTTT (\(p=0.0045\)).
To further assess robustness across different motion regimes, we repeated the same six pairwise tests independently within each motion-severity subgroup (mild, moderate, and severe), yielding \(18\) severity-stratified comparisons. 
MotionDPS achieved statistical significance in \(17/18\) tests. 
The only non-significant result corresponded to SSIM versus MotionTTT under severe motion corruption (\(p=0.1875\)).

Fig.~\ref{fig:pmoc3d_recon} presents qualitative reconstruction results for subjects S2\_1 and S4\_1 from the \emph{severe} and \emph{moderate} motion categories, respectively.
The RSS reconstruction exhibits severe blurring and ghosting artifacts. Although AltOpt and MotionTTT reduce these artifacts, both methods retain noticeable residual motion effects, particularly at cortical boundaries and in fine-scale anatomical structures. 
StackedUnet produces visibly over-smoothed reconstructions with attenuation of high-frequency detail. 
In contrast, MotionDPS more effectively suppresses motion artifacts while preserving anatomical structure, which is reflected in lower-magnitude and more spatially uniform residual errors in the corresponding error maps.
Similarly, Fig.~\ref{fig:pmoc3d_recon_lower_ssim} presents two challenging cases where MotionDPS does not outperform all competing methods in terms of SSIM. 
Nevertheless, MotionDPS produces reconstructions that remain visually comparable to the best-performing baseline.
The lower SSIM can be attributed to the smoothing effect of the denoising diffusion prior, which slightly affect local texture and contrast consistency captured by SSIM.

On PMoC3D, MotionDPS requires approximately 25 minutes per volume on an NVIDIA A100 GPU, with a peak GPU memory consumption of approximately 46 GB during inference.
In comparison, StackedUnet, AltOpt and MotionTTT require approximately 5 seconds, 4 hours and 1 hour per volume, respectively. 
These results demonstrate a favorable accuracy--runtime trade-off of MotionDPS among the physics-based baselines. Overall, MotionDPS improves robustness to real-world motion corruption, reduces structured ghosting, and restores sharp tissue boundaries that are critical for preserving diagnostic features in 3D neuroimaging.

\subsection{Ablation Studies}
 \label{sec:ablations}
We perform a series of ablation studies to analyze the contribution of the main components of MotionDPS and to better understand the factors driving its performance. 
In particular, we investigate the role of the diffusion image prior, joint coil sensitivity estimation, motion regularization, adaptive preconditioning, the number of reverse diffusion iterations, and the sensitivity of the method to its main hyperparameters.
All ablation experiments are conducted on motion-simulated FastMRI data with acceleration factor \(R=4\), which enables quantitative evaluation of both reconstruction quality and motion estimation accuracy using ground-truth motion trajectories. 
Motion corruption follows the same simulation protocol and severity levels described in Sec.~\ref{subsec:motion_sim}. 

\subsubsection{Image prior}
We compare different diffusion prior formulations, including (i) a complex-valued prior trained via two-stage learning (FastMRI pretraining + CC359 fine-tuning), (ii) a complex-valued prior trained on CC359 only, and (iii) a real-valued prior. 
The real-valued prior is trained following the same procedure described in Sec.~\ref{sec:training_details}, but on magnitude images only, and is applied independently to the real and imaginary components during inference, thereby neglecting their intrinsic coupling.
In contrast, the complex-valued priors operate directly in \(\C\), which is essential for phase estimation.

\begin{figure}[t]
    \centering
    \includegraphics[width=1.0\linewidth]{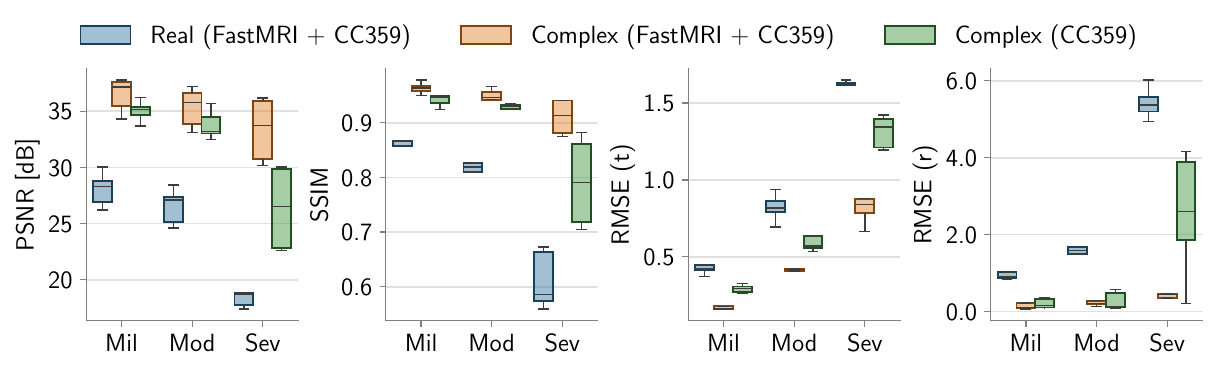}
    \caption{Comparison of diffusion prior formulations. We evaluate a real-valued prior (trained on magnitude images) and complex-valued priors with and without two-stage training.}
    \label{fig:real_cpx_prior}
\end{figure}

As shown in Fig.~\ref{fig:real_cpx_prior}, the complex-valued prior consistently improves the results, with additional gains from the two-stage training strategy. 
These observations highlight the importance of both modeling complex-valued MR signals and leveraging large-scale pretraining for capturing image statistics.

\subsubsection{Joint coil estimation}
\label{subsec:coil_ablation}
Accurate coil sensitivity estimation is critical for multi-coil MRI reconstruction, as errors in the coil maps directly propagate into the reconstructed image and can bias motion estimation. 
Most existing motion-correction pipelines rely on ESPIRiT calibration from the acquired k-space data~\cite{cordero2016, klug2024, almasni2021}, which implicitly assumes a static object during acquisition. 
In the presence of motion, however, this assumption is violated, potentially leading to corrupted sensitivity estimates and degraded reconstruction quality.
To evaluate this effect, we compare joint coil estimation against fixed ESPIRiT estimates from motion-free k-space (reference setting) and motion-corrupted k-space (practical setting).

\begin{figure}[t]
     \centering
    \includegraphics[width=1.0\linewidth]{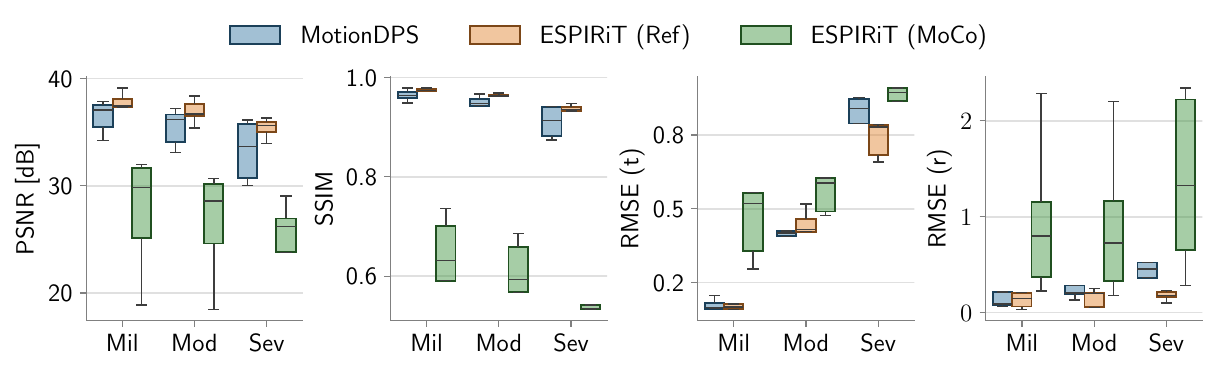}
     \caption{Comparison of different coils estimation methods. We compare joint coil estimation (MotionDPS) with ESPIRiT coils estimated from motion-free k-space (Ref) and motion-corrupted k-space (MoCo).}
     \label{fig:coil_ablation}
 \end{figure}

Fig.~\ref{fig:coil_ablation} highlights that MotionDPS achieves performance comparable to the reference setting, whereas ESPIRiT coils estimated from motion-corrupted k-space produce unreliable sensitivity estimates, degrading image quality and motion estimation accuracy even under mild motion. 
This behavior motivates the joint estimation of coil sensitivities within the reconstruction algorithm.

 \subsubsection{Motion regularization and preconditioning}
We compare the full MotionDPS model against variants without second-order motion regularization, without the diagonal preconditioner, and without both components. 

\begin{figure}[t]
    \centering
    \includegraphics[width=1.0\linewidth]{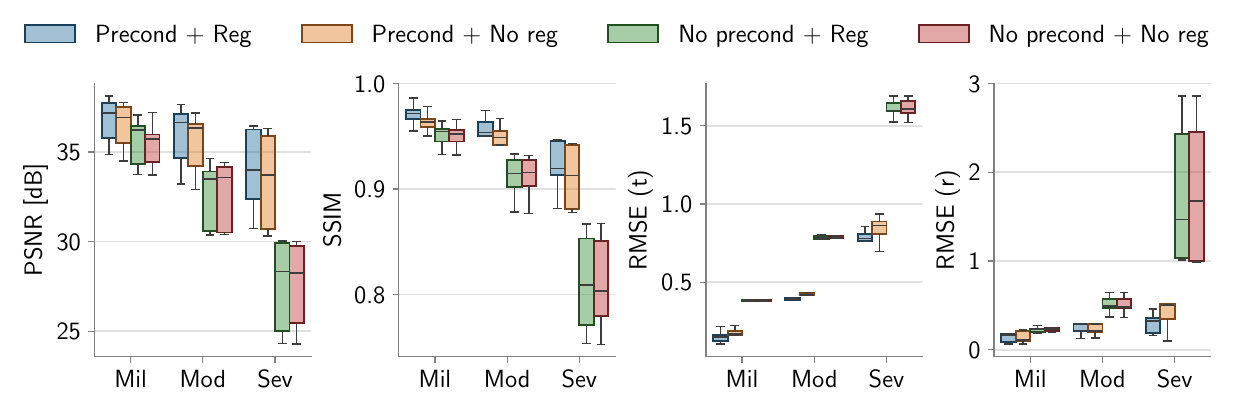}
    \caption{Ablation of temporal motion regularization and adaptive preconditioning.}
    \label{fig:motion_precond_ablation}
\end{figure}

As shown in Fig.~\ref{fig:motion_precond_ablation}, adaptive preconditioning provides the primary performance gain, and its impact becomes significant as motion severity increases.
In contrast, temporal regularization yields smaller gains by enforcing smooth trajectories. 
The combination of both achieves the best performance, while removing either degrades results, highlighting their complementary roles in better conditioning the motion optimization subproblem.
In particular, we found temporal motion regularization to be especially important under severe motion.

\subsubsection{Sensitivity to reverse diffusion iterations}
\begin{figure}[t]
    \centering
    \includegraphics[width=1.0\linewidth]{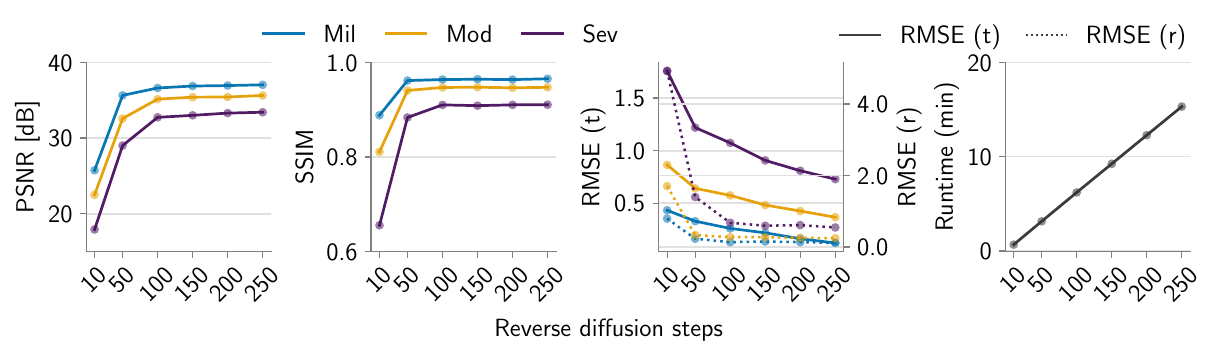}
    \caption{Effect of the number of reverse diffusion iterations on reconstruction performance, motion estimation accuracy, and computational cost.}
    \label{fig:diffusion_steps}
\end{figure}

We investigate the sensitivity of MotionDPS with respect to the number of reverse diffusion iterations \(N\) and its impact on reconstruction quality and motion estimation accuracy.
As shown in Fig.~\ref{fig:diffusion_steps}, the required number of iterations depends on the severity of motion. 
For mild motion, satisfactory reconstructions are achieved with as few as \(N=50\) iterations. 
For moderate motion, performance stabilizes around \(N=100\), while severe motion requires a larger number of iterations (\(N\approx200\)) to achieve high-quality reconstructions.
Additionally, the overall runtime scales approximately linearly with \(N\), confirming that MotionDPS provides a direct trade-off between reconstruction accuracy and computational cost.

We further analyze the computational cost of the different components of MotionDPS. 
On average, the image update accounts for approximately \(23\%\) of the total runtime, the coil sensitivity update for \(30\%\), and the motion estimation step for \(47\%\). 
This indicates that the motion subproblem constitutes the primary computational bottleneck of MotionDPS.

\subsubsection{Hyperparameters}
We analyze the sensitivity of MotionDPS with respect to its key hyperparameters:
the coil regularization weight \(\gamma\), the motion regularization weight \(\eta\), and the DPS data-consistency step size \(\zeta\).

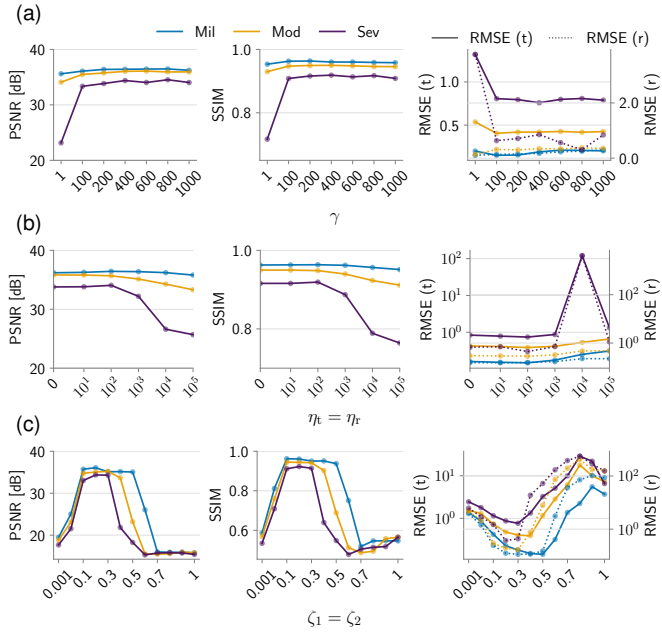
\begin{figure}[t]
\centering
\begin{tikzpicture}

\node[inner sep=0pt] (a) at (0,0) {
    \includegraphics[width=1.0\linewidth]{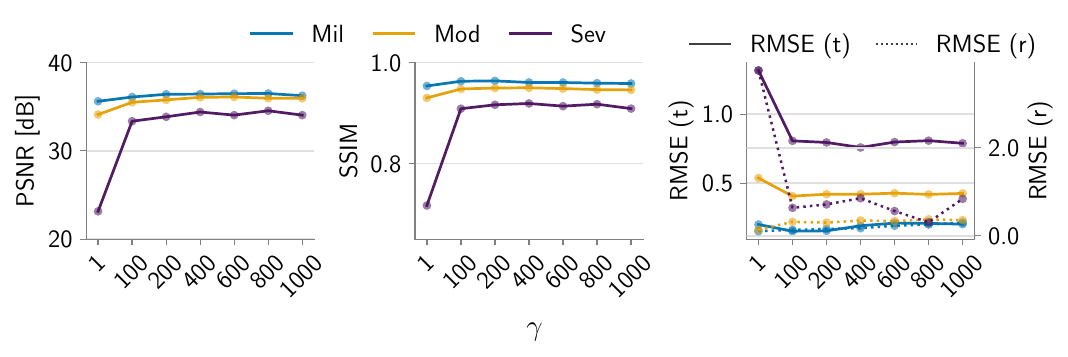}
};

\node[anchor=north west, font=\footnotesize\sffamily] at ($(a.north west)+(0.1,0.2)$) {(a)};

\node[below=0.01cm of a, inner sep=0pt] (b) {
    \includegraphics[width=1.0\linewidth]{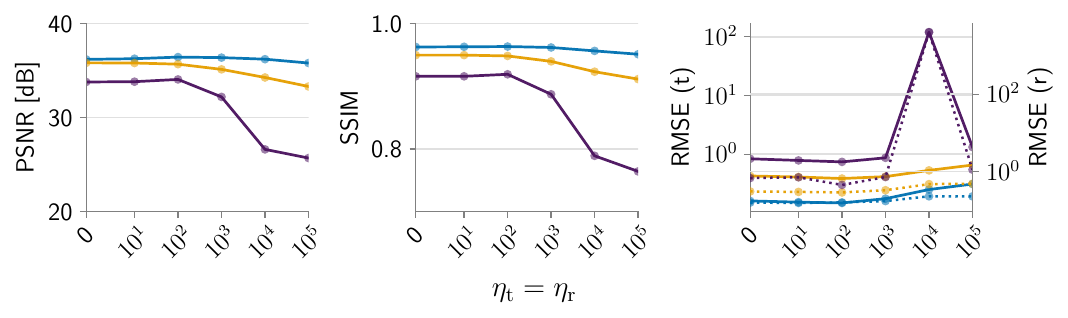}
};

\node[anchor=north west, font=\sffamily\footnotesize] at ($(b.north west)+(0.1,0.4)$) {(b)};

\node[below=0.01cm of b, inner sep=0pt] (c) {
    \includegraphics[width=1.0\linewidth]{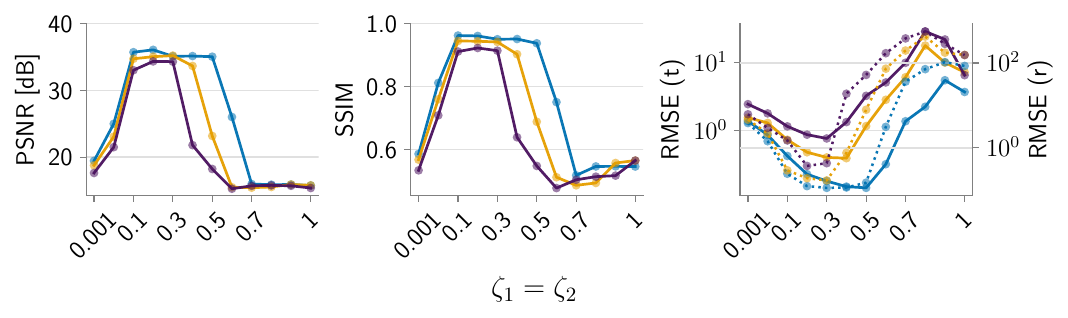}
};

\node[anchor=north west, font=\sffamily\footnotesize] at ($(c.north west)+(0.1,0.4)$) {(c)};

\end{tikzpicture}

\caption{Hyperparameter sensitivity analysis of MotionDPS. 
(a) Effect of the coil sensitivity regularization weight \(\gamma\). 
(b) Effect of the motion regularization weight \(\eta\). 
(c) Effect of the DPS data-consistency step size \(\zeta\).}
\label{fig:regularization_ablation}
\end{figure}

The plots in Fig.~\ref{fig:regularization_ablation} demonstrate that MotionDPS remains stable across a broad range of hyperparameter values. 
Small \(\gamma\) values produce unstable coil estimates, whereas excessively large values oversmooth the sensitivities. 
Moderate \(\eta\) values improve motion stability, while strong regularization increases motion RMSE due to oversmoothing of the trajectory. 
Among all parameters, \(\zeta\) has the strongest effect: small values weaken data consistency, whereas large values destabilize the reverse diffusion process. 
We observe similar trends on CC359 and PMoC3D, indicating that MotionDPS is reasonably robust across datasets and acquisition settings. 
Empirically, moderate-to-large coil regularization \(\gamma \geq 100\), motion regularization in \(0 < \eta \leq 1000\), and DPS step sizes in \(0.1 \leq \zeta \leq 0.4\) provide an initial reliable operating regimes for new datasets.

\section{Discussion}
\label{sec:discussion}

This work introduces MotionDPS, a diffusion-based framework for motion-compensated 3D multi-coil MRI reconstruction that jointly estimates the motion-free anatomical image, rigid-body motion trajectory, and coil sensitivity profiles from accelerated, motion-corrupted k-space data. 
Our main contribution lies in integrating a score-based diffusion prior into a physics-consistent variational formulation, where posterior inference is carried out via deterministic reverse-diffusion sampling coupled with proximal updates for motion and coil variables. 
This design enables fully unsupervised reconstruction, without requiring paired motion-free and motion-corrupted training data or external motion tracking devices.

Across both simulated and real-motion experiments, MotionDPS demonstrates strong robustness to motion corruption with and without undersampling. 
In simulation, reconstruction quality degrades with increasing motion severity and acceleration factor, while maintaining high PSNR and SSIM values even under severe corruption. Moreover, the recovered motion trajectories closely match the ground-truth trajectories.

In real-motion experiments, MotionDPS achieves the highest PSNR and SSIM across \emph{all} motion categories and exhibits improved structural fidelity in qualitative comparisons. 
In particular, MotionDPS better preserves fine anatomical structures and sharp tissue boundaries, highlighting that combining an explicit physics-based forward model with an expressive diffusion prior provides effective guidance toward anatomically plausible yet data-consistent reconstructions.
However, the real-data evaluation remains limited to adult brain MRI from PMoC3D. The dataset contains a small number of subjects and does not include multiple scanner vendors, anatomies, pathologies, or pediatric cohorts. Future work include the evaluation on larger datasets and reader studies to assess diagnostic relevance.

Despite these advantages, some limitations remain. 
First, the current formulation assumes a rigid-body motion model, which may not adequately capture more complex non-rigid motion patterns encountered in 3D brain imaging, such as respiration-induced deformation, eye or tongue movements, or brain pulsation. 
Importantly, MotionDPS is not fundamentally restricted to rigid motion, as the motion operator \(\mathcal{\oW}\) can be extended to parameterize more flexible deformation fields.
Such an extension would require modifications to the motion prior and the optimization scheme to account for the increased model complexity.
Therefore, extending the framework to incorporate deformable motion models represents an important direction for future work.
Second, MotionDPS requires repeated evaluation and backpropagation through a 3D diffusion denoiser during inference, in addition to inner-loop proximal updates for motion states and coil sensitivities, resulting in high computational cost.
Although MotionDPS achieves the lowest runtime among the physics-based baselines in our experiments, reducing the number of network evaluations remains an important direction for future work.

A further important direction for future work is the evaluation of MotionDPS in pediatric imaging scenarios, where patient motion is typically more frequent, unpredictable, and pronounced.
Moreover, the extension to low-field MRI systems would be important as these devices become increasingly widespread due to reduced cost and improved accessibility in resource-limited settings~\cite{Kofler2025}. 
However, low-field MRI inherently suffers from lower signal-to-noise ratio and increased sensitivity to motion, making robust motion compensation especially critical in these emerging clinical applications.
Addressing these scenarios would require adapting or fine-tuning the diffusion prior to account for differences in image contrast, noise characteristics, and anatomical variability.

\section{Conclusion}
\label{sec:conclusion}

We introduced MotionDPS, a novel method that integrates diffusion-based image priors with explicit motion-aware data consistency to enable  joint image and coil reconstruction as well as motion estimation for 3D MRI.
To the best of our knowledge, this is the first method to jointly estimate the image, coil sensitivity maps, and motion states in a fully 3D MRI setting.
Across both simulated and real-motion datasets, the method demonstrates superior robustness to motion and undersampling, consistently achieving higher PSNR and SSIM, generating cleaner reconstructions, while at the same time reducing runtime compared to competitive baselines.
These results support MotionDPS as a promising framework for 3D brain MRI reconstruction in the presence of patient motion.

\bibliographystyle{IEEEtran}
\bibliography{references}

\end{document}